%% file: TriDimEEG.tex
\documentclass[10pt,letterpaper,twocolumn]{article}
\usepackage[T1]{fontenc}
\usepackage{newtxtext}
\usepackage[margin=1in,includefoot,columnsep=0.3in]{geometry}
\usepackage[table]{xcolor}
\usepackage{graphicx}
\usepackage{amsmath,amssymb}
\usepackage[authoryear,round]{natbib}
\usepackage{caption}
\usepackage{booktabs}
\usepackage{multirow}
\usepackage{array}
\usepackage{tabularx}
\usepackage{microtype}
\usepackage{placeins}
\usepackage{flushend}
\usepackage{titlesec}
\titleformat{\section}{\normalfont\fontsize{12}{14}\selectfont\bfseries\centering}{\thesection}{0.6em}{}
\titleformat{\subsection}{\normalfont\fontsize{10}{12}\selectfont\bfseries\centering}{\thesubsection}{0.6em}{}
\titleformat{\subsubsection}{\normalfont\normalsize\bfseries\centering}{\thesubsubsection}{0.6em}{}
\titlespacing*{\section}{0pt}{13pt plus 2pt minus 2pt}{6pt}
\titlespacing*{\subsection}{0pt}{10pt plus 2pt minus 2pt}{4pt}
\titlespacing*{\subsubsection}{0pt}{8pt plus 2pt minus 2pt}{4pt}
\usepackage{xurl}
\definecolor{linkblue}{RGB}{35, 80, 130}
\usepackage[colorlinks=true,linkcolor=black,citecolor=black,urlcolor=linkblue]{hyperref}
\makeatletter
\renewcommand{\maketitle}{%
    \begin{center}
        {\fontsize{14}{17}\selectfont\bfseries\@title\par}
        \vspace{10pt}
        \@author
    \end{center}
}
\makeatother
\renewenvironment{abstract}{%
    \normalsize
    \setlength{\parindent}{0pt}
    \setlength{\parskip}{4pt}
    {\centering\fontsize{11}{13}\selectfont\bfseries\abstractname\par}
    \vspace{5pt}\noindent
}{\par}

\title{Beyond Flattened Tokens: Structure-Preserving EEG Decoding\\with Reusable TriDim Blocks}
\author{
    \begin{minipage}{\textwidth}
    \centering
    \fontsize{10}{13}\selectfont
    Shiyue Su\textsuperscript{\rm 1,\rm 2},
    Song Wang\textsuperscript{\rm 2},
    Zekai Zhan\textsuperscript{\rm 1},
    Junjie Zeng\textsuperscript{\rm 1},
    Ziling Lu\textsuperscript{\rm 1},\\
    Zongsheng Li\textsuperscript{\rm 1,\rm 4},
    Xinyuan Ye\textsuperscript{\rm 1},
    Zhiyuan Ma\textsuperscript{\rm 5},
    Xinke Shen\textsuperscript{\rm 1*}, and
    Quanying Liu\textsuperscript{\rm 1,\rm 2,\rm 3*}\\[8pt]
    \fontsize{10}{12}\selectfont
    \textsuperscript{\rm 1}Department of Biomedical Engineering, Southern University of Science and Technology,\\
    Shenzhen, 518055, China.\\
    \textsuperscript{\rm 2}Omni-Intelligence, Shenzhen, China.\\
    \textsuperscript{\rm 3}Shenzhen Loop Area Institute, Shenzhen, China.\\
    \textsuperscript{\rm 4}Department of Computer Science, The Chinese University of Hong Kong,\\
    Shenzhen, 518172, China.\\
    \textsuperscript{\rm 5}School of Biomedical Engineering, Tsinghua Medicine, Tsinghua University,\\
    Beijing, 100084, China.\\
    \texttt{shenxk@sustech.edu.cn; liuqy@sustech.edu.cn}\\
    \textsuperscript{*}Co-corresponding authors
    \end{minipage}
}
\date{}

\begin{document}

\twocolumn[{
\begin{minipage}{\textwidth}
\maketitle

\begin{abstract}
Effective EEG decoding requires representations that preserve organization
among channels, local waveform dynamics, and long-range temporal context.
Existing EEG architectures often capture these structures using separate
specialized modules or collapse them into a single token sequence, making it
difficult to maintain their distinct roles and coordinate their interactions
throughout the backbone. We propose TriDim, a reusable block that preserves
the representation shape and keeps three EEG axes explicit: channel, sample
position within each patch, and patch position across the recording. These
axes correspond to spatial, short-term temporal, and long-term temporal
information, respectively. Each TriDim block applies feed-forward
transformations along individual axes and cross-axis attention to coordinate
information exchange among them. By stacking TriDim blocks with a multi-level
tri-axis readout, we construct TriDimEEG, a standalone EEG decoder. Under
strict cross-subject evaluation on eight datasets spanning clinical
diagnosis, sleep staging, motor imagery, and emotion recognition, TriDimEEG
achieves the best overall performance among fifteen evaluated models, with
a 4.3\% relative improvement in average accuracy over the second-best model. Replacing Transformer
blocks in three EEG foundation models with TriDim blocks yields an average
relative improvement of 7.4\% in downstream accuracy while reducing parameter
counts by 17.0\% to 47.3\%. These results establish TriDim as an effective and
reusable building block and TriDimEEG as a strong standalone EEG decoder.

Code and parameters of TriDimEEG are available at
\url{https://github.com/ncclab-sustech/TriDim_model}.

\end{abstract}
\vspace{14pt}
\end{minipage}
}]

\section{Introduction}

\begin{figure}[t]
  \centering
  \IfFileExists{Figures/3_information.jpg}{%
    \includegraphics[width=0.9\linewidth]{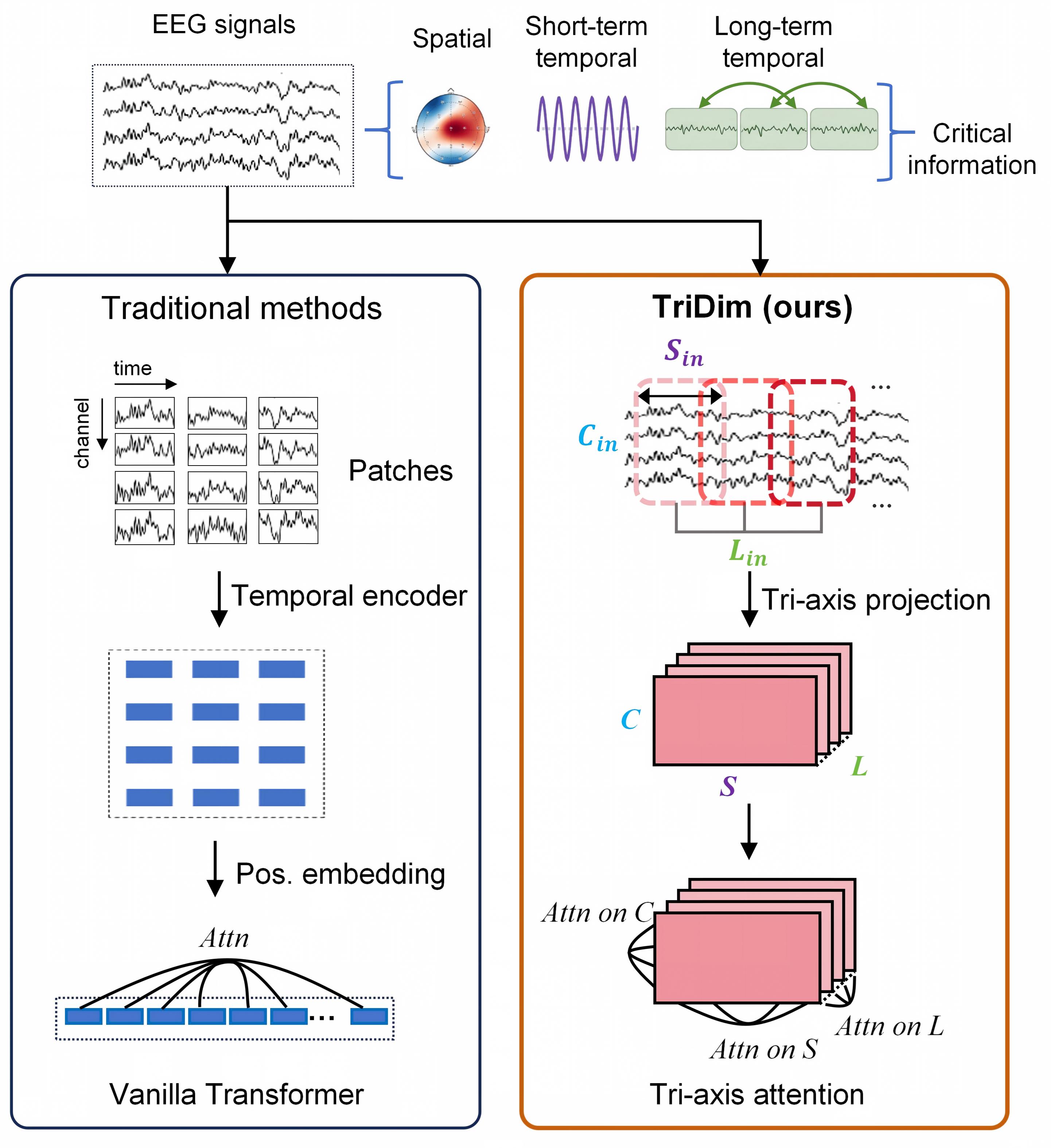}%
  }{%
    \fbox{\parbox{0.9\linewidth}{\centering Missing figure:
    \texttt{\detokenize{Figures/3_information.jpg}}}}%
  }
  \caption{Comparison of conventional flattened-token EEG architecture and
TriDim. TriDim preserves channel, within-patch temporal, and across-patch
temporal structures as three explicit axes for tri-axis modeling.}
  \label{fig:illustration}
\end{figure}

Electroencephalography (EEG) decoding supports diverse applications,
such as clinical diagnosis, sleep monitoring, robotic control, and
affective computing~\citep{craik2019deep,roy2019deep}. EEG signals can be characterized by three complementary structural aspects (Figure~\ref{fig:illustration}): spatial organization across channels,
waveform dynamics within local windows, and long-range temporal context
across windows~\citep{lawhern2018eegnet,ding2022tsception,
song2022eeg,wang2024cbramod}. Spatial organization captures how neural activity is
distributed and coordinated across recording channels. Local waveform
dynamics characterize oscillatory and transient patterns within short
intervals. Long-range temporal context describes how these patterns evolve
across extended periods. The three information types are also strongly
interdependent~\citep{wang2024cbramod,wang2026brastorm}. For example, the same local waveform may indicate different
neural states when expressed over different channel distributions, and the
meaning of a spatial pattern may change according to the preceding and
subsequent temporal context. Effective EEG decoding therefore requires a backbone that progressively
refines each information type and repeatedly coordinates interactions among the three information types. Existing architectures, however, often
prioritize only part of this structure and do not provide sufficiently deep
and coordinated modeling of all three information types.


The main limitation lies in how current EEG architectures organize
computation across the spatial and temporal dimensions of EEG. Compact
convolutional networks use temporal kernels and spatial filters to capture
local dynamics and channel interactions
\citep{lawhern2018eegnet,ding2022tsception,miao2023lmda}, but predefined
receptive fields constrain their ability to model long-range dependencies.
Transformer-based models address this limitation in different ways.
CNN--Transformer hybrids combine convolutional feature extraction with
self-attention
\citep{song2022eeg,wan2023eegformer,ding2024eeg}. Many pretrained EEG encoders divide multichannel recordings into
channel-wise temporal patches and arrange the resulting channel--time
patches as a single token sequence for Transformer processing
\citep{yang2023biot,jiang2024large,wang2024eegpt}. As illustrated in Figure~\ref{fig:illustration}, this conventional
flattened-token pipeline mixes spatial and temporal structures within a
single sequence-processing pathway. Unlike language, EEG is
a continuous multichannel signal with
temporal structure at multiple scales. A single token sequence assigns
spatial organization and temporal dependencies to the same sequence-processing pathway. This design limits the depth and specificity with which the spatial, local temporal, and global temporal information and their interactions can be modeled. Criss--cross Transformers partly preserve EEG structure by applying
self-attention along separate spatial and temporal directions
\citep{wang2024cbramod,wang2026brastorm}. However, these models still assign
fixed computational roles to the three EEG axes. Channel and
across-patch time are treated as attention axes, whereas ordered within-patch
samples remain in the embedding dimension. Consequently, within-patch
representations are repeatedly refined by feed-forward networks without explicit relation modeling through self-attention. Channel and across-patch
representations are repeatedly mixed through attention without axis-wise
feed-forward transformation. Because self-attention and feed-forward networks
provide complementary relational modeling and nonlinear representation
refinement \citep{dong2021attention,geva2020transformer,xu2024enhancing}, this fixed assignment limits the deep processing of the three
information types and the progressive refinement of their interactions
throughout the backbone.

\begin{figure*}[t]
\centering
\IfFileExists{Figures/workflow.jpg}{%
\includegraphics[width=\textwidth]{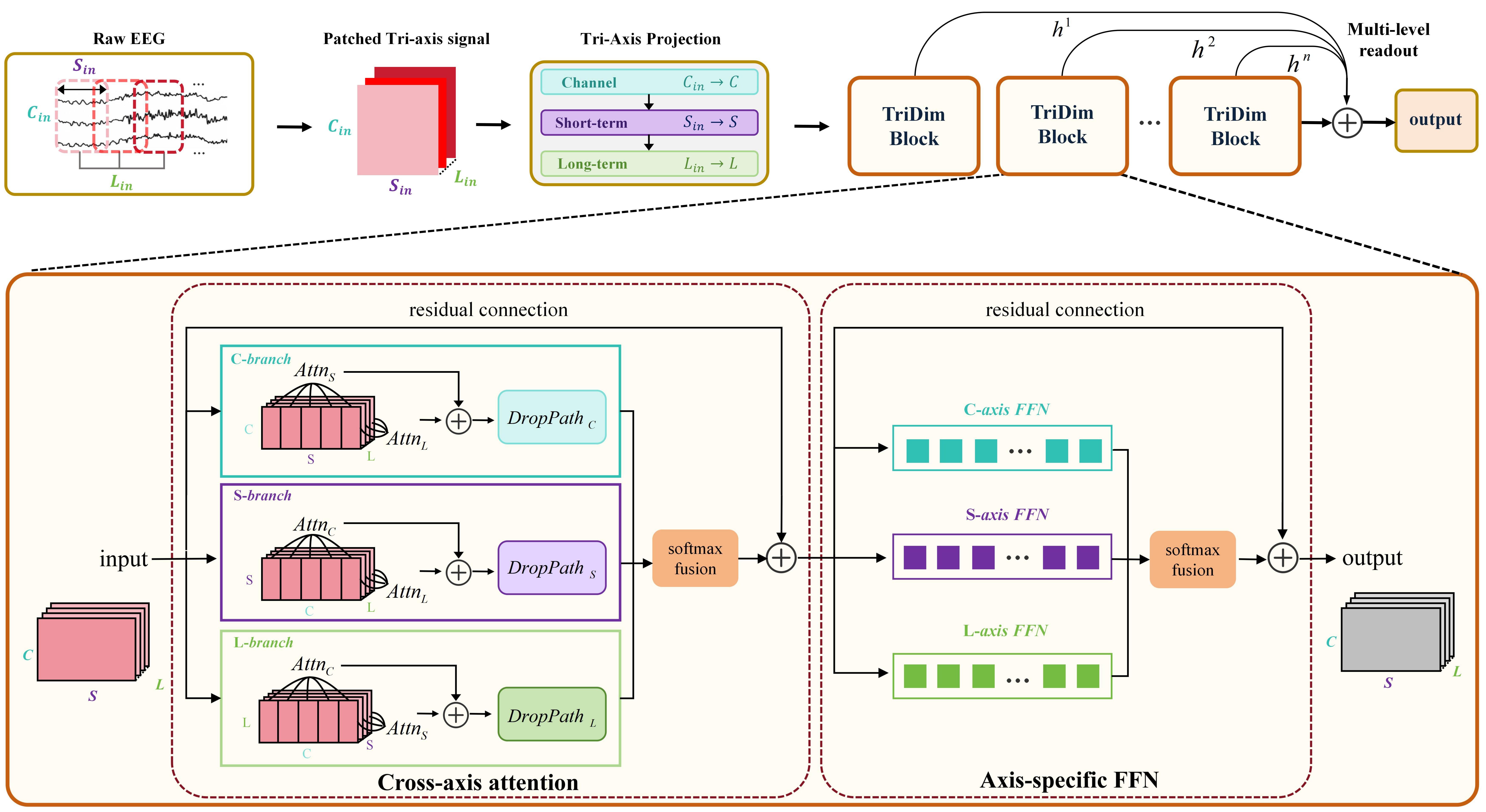}%
}{%
\fbox{\parbox{0.9\textwidth}{\centering Missing figure:
\texttt{Figures/workflow.jpg}}}%
}
\caption{Overview of TriDimEEG, including tri-axis input projections,
stacked TriDim blocks, and a multi-level tri-axis readout.}
\label{fig:workflow}
\end{figure*}

To enable deep processing of all three EEG information types and their interactions, we adopt a different design principle: channel, within-patch time, and
across-patch time should remain explicit representation axes throughout the
backbone. Based on this
principle, we propose TriDim, a shape-preserving block for EEG modeling
(Figure~\ref{fig:workflow}). TriDim constructs three axis-conditioned views. In
each view, one axis serves as the embedding dimension, and self-attention is
applied along the other two axes. The resulting views are combined through
learnable weighted aggregation. Stacking TriDim blocks therefore enables spatial,
short-term temporal, and long-term temporal information to undergo repeated
feed-forward transformation, and interactions among the three information types
are progressively updated through self-attention. A multi-level tri-axis readout
further aggregates representations from different encoder depths. The
resulting TriDimEEG model can serve as a task-specific EEG decoder, and the
same TriDim block can also replace Transformer blocks in pretrained EEG encoders without redesigning the surrounding pipelines. This work makes three main contributions:

\begin{itemize}
    \item We reinterpret temporally patched EEG representations using three
    explicit axes corresponding to channel, sample position within each
    patch, and patch position along the recording. This organization
    preserves spatial, short-term temporal, and long-term temporal
    information throughout the backbone.

    \item We develop TriDim, a reusable block that combines feed-forward
    transformations along individual axes with cross-axis attention. We
    further construct TriDimEEG by stacking TriDim blocks and introducing
    a multi-level tri-axis readout that aggregates representations across
    encoder depths.

    \item Strict cross-subject experiments on eight EEG benchmarks establish
    TriDimEEG as a strong standalone decoder. Replacing Transformer blocks
    in three EEG foundation models further demonstrates the reusability
    and parameter efficiency of TriDim.
\end{itemize}

\section{Related Work}

\subsection{Task-Specific EEG Decoders}

Early EEG decoders primarily relied on convolutional networks to
capture local waveform patterns and inter-channel relationships. EEGNet
combines temporal convolution with depthwise spatial filtering in a compact
architecture~\citep{lawhern2018eegnet}. Subsequent supervised models
introduce multi-scale temporal kernels and specialized spatial processing for different EEG tasks~\citep{ding2022tsception,miao2023lmda,ma2026dsainet}.
Attention-based architectures further extend this paradigm by modeling
dependencies beyond local convolutional receptive fields. EEGConformer
applies self-attention to features extracted by a convolutional
temporal--spatial encoder~\citep{song2022eeg}. EEGDeformer introduces
hierarchical coarse-to-fine Transformer processing and dense information
purification to aggregate representations across multiple levels
~\citep{ding2024eeg}. TeCh replaces conventional pairwise self-attention
with Core Token Aggregation-Redistribution, which collects global context
into a core token and redistributes it to the remaining tokens
~\citep{yu2026decentralized}. Despite their architectural differences,
these methods generally convert EEG signals into feature sequences before
applying attention-based contextual modeling.
\subsection{Pretrained EEG Representation Models}
Large-scale pretraining has further established sequence modeling as a
prominent paradigm for transferable EEG representation learning. BENDR~\citep{kostas2021bendr},
BIOT~\citep{yang2023biot}, LaBraM~\citep{jiang2024large}, and REVE~\citep{elouahidi2025reve} learn contextual representations from segmented
EEG recordings through convolutional or Transformer-based
encoders. Recent pretrained models additionally explore
language-aligned representations~\citep{cui2024neuro}, state-space sequence modeling~\citep{wang2025eegmamba},
frequency-enhanced encoders~\citep{tegon2025femba}, and sensor-aware tokenization
~\citep{xiao2025brainomni}. These approaches have substantially
improved transfer across datasets and downstream tasks. Nevertheless, despite differences in backbone architectures and pretraining objectives, most of these models still arrange local EEG segments into a one-dimensional token sequence, leaving spatial organization and across-segment temporal dependencies to be modeled within the same sequence-processing pathway.

\subsection{Structure-Preserving EEG Architectures}

Recent studies have begun to design attention mechanisms more explicitly aligned with the structural organization of EEG. CBraMod introduces a criss-cross
Transformer that separately performs spatial attention across electrodes
and temporal attention across windows~\citep{wang2024cbramod}. BraSTORM
adopts separate spatial and temporal branches to learn complementary
representations through input-based spatiotemporal decomposition
~\citep{wang2026brastorm}. CSBrain further introduces cross-scale
spatiotemporal tokenization and structured sparse attention to model
dependencies across temporal windows and anatomical regions
~\citep{zhou2025csbrain}. These methods demonstrate the value of preserving
structural distinctions in EEG rather than treating all signal components
as positions in a homogeneous sequence. However, existing structure-aware
architectures still assign fixed and asymmetric computational roles to the
EEG axes, typically treating channel and across-patch time as attention axes
and retaining within-patch samples as the embedding dimension.
TriDim removes this fixed assignment by preserving channel, within-patch
time, and across-patch time as three explicit axes, allowing each axis to
undergo both axis-specific feed-forward transformation and cross-axis
attention throughout the backbone.


\section{Methods}

TriDimEEG consists of a tri-axis projection, an
encoder constructed from $N$ stacked TriDim blocks, and a multi-level
tri-axis readout.

\subsection{Tri-axis Representation}

We organize EEG representations along three axes: channel organization,
short-term temporal structure, and long-term temporal context. Given an
input batch
$X_{\mathrm{raw}}\in
\mathbb{R}^{B\times C_{\mathrm{in}}\times T_{\mathrm{raw}}}$
with $B$ EEG samples, $C_{\mathrm{in}}$ electrodes, and $T_{\mathrm{raw}}$ time points, we first apply
$\mathrm{InstanceTimeNorm}$ independently to each channel over time. This
operation removes the temporal mean and normalizes the amplitude scale of
each channel without changing the tensor shape:
\[
\hat{X}
\in
\mathbb{R}^{B\times C_{\mathrm{in}}\times T_{\mathrm{raw}}}.
\]

We then unfold the time axis into $L_{\mathrm{in}}$ patches of length
$S_{\mathrm{in}}$ with stride $R$, applying zero-padding at the end when
necessary. This operation separates the original temporal dimension into
the ordered samples within each patch and the ordered patch positions
along the segment:
\[
\hat{X}
\in
\mathbb{R}^{B\times C_{\mathrm{in}}\times T_{\mathrm{raw}}}
\;\longrightarrow\;
X^{\mathrm{patch}}
\in
\mathbb{R}^{B\times C_{\mathrm{in}}\times S_{\mathrm{in}}
\times L_{\mathrm{in}}}.
\]

Three basis projections, one for each representation axis, map
$(C_{\mathrm{in}},S_{\mathrm{in}},L_{\mathrm{in}})$
to the latent dimensions $(C,S,L)$:
\[
\begin{aligned}
\mathbb{R}^{C_{\mathrm{in}}}
&\to
\mathbb{R}^{C}
&& \text{(}1{\times}1\text{ convolution)},\\
\mathbb{R}^{S_{\mathrm{in}}}
&\to
\mathbb{R}^{S}
&& \text{(linear projection)},\\
\mathbb{R}^{L_{\mathrm{in}}}
&\to
\mathbb{R}^{L}
&& \text{(shared convolution and pooling)}.
\end{aligned}
\]

The channel projection applies a $1{\times}1$ convolution along the
electrode axis $C_{\mathrm{in}}$. The short-term projection applies a linear
transformation along $S_{\mathrm{in}}$. The long-term projection applies
a shared one-dimensional convolution to the patch sequence at every
channel and within-patch position, followed by GELU, dropout, and adaptive
average pooling along $L_{\mathrm{in}}$. The three projections produce
\[
X^{\mathrm{patch}}
\in
\mathbb{R}^{B\times C_{\mathrm{in}}\times S_{\mathrm{in}}
\times L_{\mathrm{in}}}
\;\longrightarrow\;
X^{0}
\in
\mathbb{R}^{B\times C\times S\times L}.
\]

We use
$\mathcal{A}=\{c,s,l\}$
to denote the latent channel, short-term, and long-term axes, whose sizes
are $C$, $S$, and $L$, respectively. The TriDimEEG encoder operates on the resulting $[B,C,S,L]$ representation, which is preserved
throughout all TriDim blocks.

\subsection{TriDim Block}

Let
$X^{\ell}\in\mathbb{R}^{B\times C\times S\times L}$
denote the input to block $\ell+1$, where
$\ell=0,\ldots,N-1$ and $X^{0}$ is defined above. Each TriDim block
contains two pre-normalized residual sublayers: a cross-axis attention
sublayer followed by feed-forward transformations along individual axes.
Both sublayers preserve the $[B,C,S,L]$ representation shape.

\textbf{Cross-axis attention.}
For each axis $a\in\mathcal{A}$, we first apply Root Mean Square
normalization~\citep{zhang2019root} along that axis:
\[
\tilde{X}^{\ell}_{a}
=
\mathrm{AxisRMSNorm}^{\mathrm{attn}}_{a}
\left(
X^{\ell}
\right),
\qquad
a\in\mathcal{A}.
\]

For a tensor $Z$, let
$\mathrm{Attn}_{p|q}(Z)$,
where $p,q\in\mathcal{A}$ and $p\neq q$, denote multi-head self-attention
along sequence axis $p$, with axis $q$ serving as the embedding dimension.
The remaining axis is treated as a collection of independent slices and
folded into the batch dimension during attention computation.

For each embedding axis $q$, the two attention operations along the
remaining axes share the same query, key, value, and output projection
parameters. They are applied independently to the same normalized
representation, and their outputs are averaged. The channel, short-term,
and long-term views are computed as
\[
\begin{aligned}
V_{c}^{\ell}
&=
\frac{1}{2}
\left[
\mathrm{Attn}_{s|c}
\left(
\tilde{X}^{\ell}_{c}
\right)
+
\mathrm{Attn}_{l|c}
\left(
\tilde{X}^{\ell}_{c}
\right)
\right],\\
V_{s}^{\ell}
&=
\frac{1}{2}
\left[
\mathrm{Attn}_{c|s}
\left(
\tilde{X}^{\ell}_{s}
\right)
+
\mathrm{Attn}_{l|s}
\left(
\tilde{X}^{\ell}_{s}
\right)
\right],\\
V_{l}^{\ell}
&=
\frac{1}{2}
\left[
\mathrm{Attn}_{c|l}
\left(
\tilde{X}^{\ell}_{l}
\right)
+
\mathrm{Attn}_{s|l}
\left(
\tilde{X}^{\ell}_{l}
\right)
\right].
\end{aligned}
\]

Each view retains the $[B,C,S,L]$ shape and is named according to its
embedding axis rather than the sequence axes over which attention is
performed.

To robustly integrate the three complementary attention views, we
regularize each view before fusion. Specifically, we apply DropPath independently to each view
$V_{a}^{\ell}$ with dropout probability $p_{a}^{\ell}$~\citep{huang2016stochastic}. This enables view-wise stochastic depth and discourages the model from relying excessively on any single EEG axis. The details are reported in Appendix~\ref{sec:hyper}.


The regularized views are then combined using learnable fusion weights. Let
\[
\boldsymbol{\alpha}^{\ell}
=
\mathrm{softmax}
\left(
\boldsymbol{\theta}_{\mathrm{attn}}^{\ell}
\right)
\in
\mathbb{R}^{3}
\]
denote the learnable view-fusion weights, and
$\gamma_{\mathrm{attn}}^{\ell}\in\mathbb{R}^{C}$ per-channel learnable LayerScale parameters~\citep{touvron2021going}.
The attention residual update is
\[
X^{\ell+\frac{1}{2}}
=
X^{\ell}
+
\gamma_{\mathrm{attn}}^{\ell}
\odot
\sum_{a\in\mathcal{A}}
\alpha_{a}^{\ell}
\,
\mathrm{DropPath}
\left(
V_{a}^{\ell},
p_{a}^{\ell}
\right).
\]

\textbf{Axis-wise feed-forward sublayer.}
After cross-axis attention with multi-view fusion, an independent feed-forward network $\mathrm{FFN}_{a}$ is applied along
each axis $a$. Each network contains two linear layers with an expansion
ratio of $2$, a GELU activation, and dropout. The intermediate
representation is independently normalized along each target axis:
\[
\begin{aligned}
\bar{X}^{\ell+\frac{1}{2}}_{a}
&=
\mathrm{AxisRMSNorm}^{\mathrm{ffn}}_{a}
\left(
X^{\ell+\frac{1}{2}}
\right),\\
U_{a}^{\ell}
&=
\mathrm{FFN}_{a}
\left(
\bar{X}^{\ell+\frac{1}{2}}_{a}
\right).
\end{aligned}
\]

The three axis-wise FFN outputs are first combined using learnable fusion
weights:
\[
\boldsymbol{\beta}^{\ell}
=
\mathrm{softmax}
\left(
\boldsymbol{\theta}_{\mathrm{ffn}}^{\ell}
\right)
\in
\mathbb{R}^{3},
\]
which denotes the relative contributions of the three axes. DropPath is applied to the fused output. With
per-channel learnable LayerScale parameters
$\gamma_{\mathrm{ffn}}^{\ell}\in\mathbb{R}^{C}$ and DropPath rate
$p_{\mathrm{ffn}}^{\ell}$, the block output is
\[
X^{\ell+1}
=
X^{\ell+\frac{1}{2}}
+
\gamma_{\mathrm{ffn}}^{\ell}
\odot
\mathrm{DropPath}
\left(
\sum_{a\in\mathcal{A}}
\beta_{a}^{\ell}U_{a}^{\ell},
p_{\mathrm{ffn}}^{\ell}
\right).
\]


\subsection{Multi-level Tri-axis Readout}

Representations from different encoder depths may contain complementary
information. We therefore attach a tri-axis attention
pooling head to every layer output $X^{\ell}$, where
$\ell\in\{1,\ldots,N\}$.

For
$X^{\ell}\in\mathbb{R}^{B\times C\times S\times L}$,
the pooling head constructs one axis-specific sequence for each axis. The
selected axis indexes the sequence positions, while the other two axes are
flattened into the feature dimension:
\[
\begin{aligned}
Z_{c}^{\ell}
&\in
\mathbb{R}^{B\times C\times SL},\\
Z_{s}^{\ell}
&\in
\mathbb{R}^{B\times S\times CL},\\
Z_{l}^{\ell}
&\in
\mathbb{R}^{B\times L\times CS},
\end{aligned}
\]
where
$SL=S\cdot L$,
$CL=C\cdot L$,
and
$CS=C\cdot S$.

An axis-specific linear projection $W_{a}^{\ell}$ maps each sequence to
a shared embedding dimension $E$, after which a learnable positional
embedding $\mathrm{PE}_{a}^{\ell}$ is added:
\[
H_{a}^{\ell}
=
Z_{a}^{\ell}W_{a}^{\ell}
+
\mathrm{PE}_{a}^{\ell}
\in
\mathbb{R}^{B\times |a|\times E},
\]
where
$|a|\in\{C,S,L\}$
is the sequence length along axis $a$.

A lightweight attention pooling operator
$\mathrm{AttnPool}_{a}^{\ell}$
assigns normalized weights to the sequence positions and aggregates them
into one $E$-dimensional vector. The three axis-specific vectors are
combined to form the representation of layer $\ell$:
\[
h^{\ell}
=
\sum_{a\in\mathcal{A}}
\omega_{a}^{\ell}
\,
\mathrm{AttnPool}_{a}^{\ell}
\left(
H_{a}^{\ell}
\right)
\in
\mathbb{R}^{B\times E},
\]
where
\[
\boldsymbol{\omega}^{\ell}
=
\mathrm{softmax}
\left(
\boldsymbol{\psi}^{\ell}
\right)
\in
\mathbb{R}^{3}
\]
are learnable fusion weights.

Finally, learnable depth-fusion weights combine the representations from
all $N$ encoder layers:
\[
\boldsymbol{\pi}
=
\mathrm{softmax}
\left(
\boldsymbol{\rho}
\right)
\in
\mathbb{R}^{N},
\qquad
z
=
\sum_{\ell=1}^{N}
\pi_{\ell}h^{\ell}
\in
\mathbb{R}^{B\times E}.
\]
The fused representation $z$ is passed to a classifier head consisting of
LayerNorm, dropout, and a linear classification layer.

\subsection{Experimental Settings}
\label{sec:experimental_settings}

\textbf{Datasets and evaluation protocols.}
We evaluate TriDimEEG on eight EEG benchmarks covering four application
scenarios: i) \textbf{Neurological disorder diagnosis}: AD65~\citep{miltiadous2023dataset}, (ii) \textbf{Sleep staging}: SleepEDF~\citep{kemp2000analysis,goldberger2000physiobank}, (iii) \textbf{Motor imagery}: BCI-IV-2A~\citep{tangermann2012review},
SHU-MI~\citep{ma2022large},
PhysioNet-MI~\citep{schalk2004bci2000,goldberger2000physiobank},
and (iv) \textbf{Emotion recognition}: FACED~\citep{chen2023large},
SEED~\citep{zheng2015investigating},
and SEED-V~\citep{liu2021comparing}.
Table~\ref{tab:dataset_statistics} summarizes the processed datasets.

We adopt a strict subject-independent evaluation protocol and divide
subjects into training, validation, and test sets with an approximately
8:1:1 ratio. All recordings, sessions, and segments from the same subject
are assigned to the same subset to prevent subject leakage. The subject-level partitioning procedure is repeated three times using different random seeds, yielding three predefined split manifests shared by all methods.

\begin{table}[t]
\centering
\normalsize
\setlength{\tabcolsep}{1.2pt}
\renewcommand{\arraystretch}{1.05}
\begin{tabular*}{\columnwidth}{@{\extracolsep{\fill}}lrrrrr@{}}
\toprule
\textbf{Dataset}
& \textbf{Subj.}
& \textbf{Segments}
& \textbf{Classes}
& \textbf{Ch.}
& \textbf{Samples} \\
\midrule
AD65         & 88  & 6,938   & 3 & 19 & 2,000 \\
SleepEDF     & 153 & 414,961 & 5 & 2  & 6,000 \\
BCI-2A       & 9   & 5,184   & 4 & 22 & 800   \\
SHU-MI       & 25  & 7,571   & 2 & 32 & 800   \\
PNet-MI      & 109 & 9,837   & 4 & 64 & 800   \\
FACED        & 123 & 10,332  & 9 & 32 & 2,000 \\
SEED         & 15  & 14,895  & 3 & 62 & 2,000 \\
SEED-V       & 16  & 11,424  & 5 & 62 & 2,000 \\
\bottomrule
\end{tabular*}
\caption{Statistics of the processed EEG datasets. BCI-2A and PNet-MI stand for BCI-IV-2A and PhysioNet-MI, respectively. Subj. and Ch. denote subjects and channels.}
\label{tab:dataset_statistics}
\end{table}

\textbf{Preprocessing.}
All recordings are resampled to 200~Hz, band-pass filtered from
0.3 to 75~Hz, and notch filtered at the power-line frequency corresponding
to each recording source. AD65, FACED, SEED, and SEED-V are divided into
10-s windows; BCI-IV-2A, SHU-MI, and PhysioNet-MI use 4-s windows; and
SleepEDF uses standard 30-s sleep epochs following OmniEEG-bench~\citep{lu2026omnieeg}.

\textbf{Baselines.}
We compare TriDimEEG with four supervised EEG architectures:
EEGNet~\citep{lawhern2018eegnet},
EEGConformer~\citep{song2022eeg},
EEGDeformer~\citep{ding2024eeg},
and TeCh~\citep{yu2026decentralized}.
We also evaluate ten pretrained EEG models:
BENDR~\citep{kostas2021bendr},
BIOT~\citep{yang2023biot},
LaBraM~\citep{jiang2024large},
CBraMod~\citep{wang2024cbramod},
NeuroGPT~\citep{cui2024neuro},
EEGMamba~\citep{wang2025eegmamba},
FEMBA~\citep{tegon2025femba},
NeuroLM~\citep{jiang2025neurolm},
BrainOmni~\citep{xiao2025brainomni},
and REVE~\citep{elouahidi2025reve}.
All methods use the same subject-level split manifests. For pretrained
models, the complete pretrained encoder is fine-tuned jointly with a task-specific
classification head on the training split. Necessary input adaptations follow the official
implementations, with further details provided in Appendix~\ref{app:baselines}.


\section{Results}

We evaluate TriDim through four analyses: cross-subject benchmarking on eight EEG datasets, integration into three pretrained EEG encoders as replacing blocks, ablation
studies, and interpretation analysis. Results are reported as
mean $\pm$ sample standard deviation over three subject-level splits, together with average rank across datasets. Macro-F1 results are provided in Appendix~\ref{app:persplit}.


\subsection{Cross-Subject EEG Decoding Performance}

Table~\ref{tab:main_results} compares TriDimEEG with representative
supervised and pretrained EEG models across clinical diagnosis, sleep
staging, motor imagery, and emotion recognition tasks.

\begin{table*}[tp]
\centering
\begingroup
\small
\setlength{\tabcolsep}{1.5pt}
\setlength{\medmuskip}{1mu}
\renewcommand{\arraystretch}{1.05}
\begin{tabular*}{\textwidth}{@{\extracolsep{\fill}}lcccccccccc@{}}
\toprule
\textbf{Model} & \textbf{AD65} & \textbf{SleepEDF} & \textbf{BCI-2A} & \textbf{SHU-MI} & \textbf{PNet-MI} & \textbf{FACED} & \textbf{SEED} & \textbf{SEED-V} & \textbf{Avg. $\uparrow$} & \textbf{Rank $\downarrow$} \\
\midrule
EEGNet & $59.6\pm12.0$ & $83.2\pm2.2$ & $32.8\pm4.2$ & $57.6\pm6.4$ & $57.5\pm7.7$ & $30.8\pm1.2$ & $45.0\pm6.4$ & $23.6\pm2.8$ & $48.76$ & $8.38$ \\
EEGConformer & $54.2\pm10.7$ & $84.0\pm0.9$ & $37.1\pm2.4$ & $56.8\pm6.8$ & $59.3\pm6.6$ & $38.7\pm0.4$ & $52.3\pm7.6$ & $25.1\pm1.9$ & $50.94$ & $6.75$ \\
EEGDeformer & $55.5\pm12.5$ & $83.1\pm1.3$ & $45.7\pm2.6$ & $\mathbf{68.7\pm13.4}$ & $\mathbf{62.5\pm8.6}$ & $45.9\pm1.1$ & $56.1\pm6.4$ & $25.2\pm2.4$ & $\underline{55.32}$ & $\underline{4.38}$ \\
TeCh & $60.1\pm8.8$ & $\underline{85.5\pm1.5}$ & $31.6\pm0.7$ & $\underline{60.5\pm7.6}$ & $56.2\pm8.1$ & $30.5\pm0.8$ & $51.5\pm1.8$ & $23.7\pm2.4$ & $49.94$ & $7.75$ \\
\midrule
BENDR & $50.1\pm4.2$ & $65.5\pm0.8$ & $25.1\pm0.2$ & $51.8\pm2.0$ & $46.6\pm3.9$ & $13.5\pm1.9$ & $39.1\pm8.2$ & $22.8\pm3.2$ & $39.32$ & $13.88$ \\
BIOT & $53.7\pm10.7$ & $20.0\pm0.0$ & $25.0\pm0.0$ & $54.0\pm2.0$ & $29.1\pm1.9$ & $18.7\pm1.1$ & $51.2\pm2.5$ & $24.0\pm3.5$ & $34.46$ & $12.88$ \\
LaBraM & $59.6\pm6.4$ & $76.0\pm1.6$ & $32.9\pm2.2$ & $57.0\pm1.3$ & $52.3\pm3.7$ & $31.7\pm1.0$ & $59.3\pm0.8$ & $\mathbf{34.6\pm8.6}$ & $50.42$ & $6.13$ \\
CBraMod & $53.8\pm5.8$ & $74.2\pm1.2$ & $\underline{46.3\pm6.8}$ & $56.9\pm0.8$ & $56.6\pm2.8$ & $\underline{51.8\pm1.4}$ & $\underline{59.7\pm2.4}$ & $27.7\pm5.0$ & $53.38$ & $5.13$ \\
NeuroGPT & $46.7\pm9.7$ & $70.6\pm0.9$ & $35.3\pm0.9$ & $53.9\pm1.5$ & $50.3\pm1.6$ & $36.0\pm1.1$ & $56.3\pm1.7$ & $26.6\pm4.3$ & $46.97$ & $9.25$ \\
EEGMamba & $50.3\pm7.3$ & $68.5\pm1.1$ & $33.0\pm0.7$ & $52.9\pm2.1$ & $32.0\pm1.2$ & $25.2\pm1.7$ & $52.5\pm3.1$ & $24.2\pm3.2$ & $42.32$ & $11.75$ \\
FEMBA & $\underline{61.2\pm14.8}$ & $74.4\pm1.8$ & $42.2\pm4.8$ & $54.5\pm0.5$ & $53.7\pm4.8$ & $28.5\pm0.5$ & $55.8\pm1.7$ & $29.3\pm5.1$ & $49.95$ & $6.88$ \\
NeuroLM & $46.0\pm5.9$ & $73.4\pm1.3$ & $37.8\pm0.4$ & $51.2\pm1.9$ & $54.0\pm3.3$ & $49.0\pm4.1$ & $57.5\pm3.2$ & $30.7\pm9.8$ & $49.95$ & $7.88$ \\
BrainOmni & $45.4\pm4.9$ & $70.5\pm2.1$ & $38.3\pm4.1$ & $55.5\pm0.5$ & $36.7\pm2.3$ & $33.9\pm2.6$ & $49.4\pm3.1$ & $26.6\pm2.1$ & $44.52$ & $10.38$ \\
REVE & $51.3\pm13.7$ & $73.1\pm2.3$ & $40.2\pm2.9$ & $58.7\pm2.5$ & $56.7\pm3.5$ & $30.7\pm2.3$ & $\mathbf{64.5\pm6.5}$ & $\underline{32.5\pm5.5}$ & $50.96$ & $5.88$ \\
\midrule
\textbf{TriDimEEG} & $\mathbf{63.8\pm6.1}$ & $\mathbf{85.6\pm2.2}$ & $\mathbf{56.6\pm2.7}$ & $60.2\pm7.2$ & $\underline{60.4\pm8.0}$ & $\mathbf{53.0\pm2.9}$ & $52.9\pm2.2$ & $29.0\pm4.3$ & $\mathbf{57.70}$ & $\mathbf{2.75}$ \\
\bottomrule
\end{tabular*}
\par
\endgroup
\caption{Cross-subject classification accuracy (\%) on eight EEG datasets. BCI-2A and PNet-MI stand for BCI-IV-2A and PhysioNet-MI, respectively.
The best and second-best results in each column are highlighted in bold and
underlined, respectively. Lower average rank indicates better overall
performance. Results are reported as mean $\pm$ standard deviation. Averages and ranks are computed across all eight datasets.}
\label{tab:main_results}
\end{table*}

TriDimEEG achieves the highest average accuracy of $57.70\%$ and the best
average rank of $2.75$, achieving a $4.3\%$ relative improvement in average
accuracy over the second-best model, EEGDeformer, and improving the average
rank from $4.38$ to $2.75$. It ranks first
on AD65, SleepEDF, BCI-IV-2A, and FACED, second on PhysioNet-MI, and third
on SHU-MI. EEGDeformer remains strongest on SHU-MI and PhysioNet-MI, whereas REVE and
LaBraM lead on SEED and SEED-V. However, these models are less consistent
across the full benchmark. Although trained without large-scale EEG
pretraining, TriDimEEG exceeds every evaluated pretrained encoder in both
average accuracy and average rank.

\subsection{Replacing Transformer Blocks in Pretrained EEG Encoders}

To evaluate TriDim as a reusable architectural block, we replace the
original Transformer-based blocks in REVE, CBraMod, and
CSBrain~\citep{zhou2025csbrain} with TriDim blocks. For each encoder, the
original and TriDim-based variants are pretrained on the same $2{,}000$ TUH
samples using identical pretraining and downstream settings. Details are provided in Appendix~\ref{app:plugin}.

\begin{figure}[t]
\centering
\IfFileExists{Figures/jizuo_2.png}{%
    \includegraphics[width=0.96\linewidth]
    {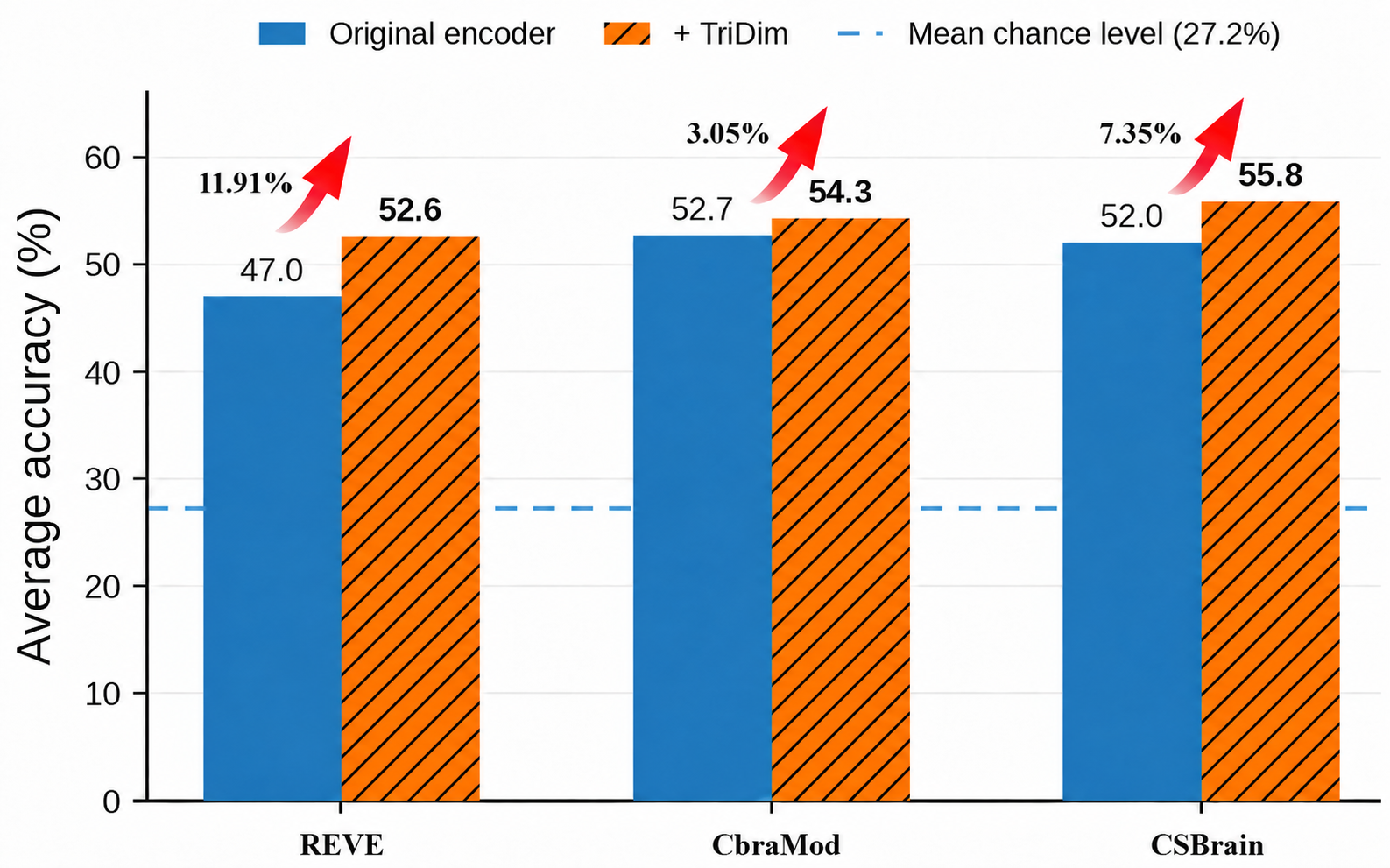}%
}{%
    \fbox{%
        \parbox[c][0.23\textheight][c]{0.94\linewidth}{%
            \centering
            \textbf{Pretrained Encoder Integration Results}\\[5pt]
            REVE, CBraMod, and CSBrain\\
            Original backbone versus backbone with TriDim blocks\\[4pt]
            Accuracy improvement and parameter reduction
        }%
    }%
}
\caption{Replacing Transformer blocks with TriDim blocks under the same
TUH-2K pretraining setting improves accuracy. Arrow annotations indicate
relative improvements over the corresponding original encoder.}
\label{fig:plugin}
\end{figure}

Replacing Transformer-based blocks with TriDim blocks yields relative
improvements in average downstream accuracy of $11.91\%$, $3.05\%$, and
$7.35\%$ over the original REVE, CBraMod, and CSBrain encoders, while
reducing parameters by $32.0\%$, $17.0\%$, and $47.3\%$, respectively.
The mean of these three relative improvements is $7.4\%$, showing that
TriDim is reusable across pretrained architectures while improving parameter
efficiency.

\subsection{Ablation Studies}

We evaluate the contributions of two aspects of TriDimEEG: the three representation axes and the design choices within
the TriDim backbone (Table~\ref{tab:ablation}). The first group removes or isolates the channel,
short-term temporal, and long-term temporal axes. The second group evaluates the multi-level readout, cross-axis attention, axis-specific feed-forward networks, attention sharing, and parallel axis integration.

Removing the
channel, short-term temporal, or long-term temporal axis decreases average
accuracy by $2.3\%$, $0.5\%$, and $1.6\%$, respectively, relative to the
Full model. The larger losses
for $C$ and $L$ highlight the importance of using channel or long-term temporal information as the embedding dimensions, which were ignored in previous studies. Retaining only one-axis branch obtain $55.44\%$, $55.04\%$, and $54.46\%$
for $C$, $S$, and $L$, respectively, all worse than the two-axis model or full TriDimEEG. Together, these
results indicate three-axis representations provide complementary information.


\begin{table*}[tp]
\centering
\begingroup
\small
\setlength{\tabcolsep}{1.5pt}
\setlength{\medmuskip}{1mu}
\renewcommand{\arraystretch}{1.05}
\begin{tabular*}{\textwidth}{@{\extracolsep{\fill}}lccccccccc@{}}
\toprule
\textbf{Variant} 
& \textbf{AD65} 
& \textbf{SleepEDF} 
& \textbf{BCI-2A} 
& \textbf{SHU-MI} 
& \textbf{PNet-MI} 
& \textbf{FACED} 
& \textbf{SEED} 
& \textbf{SEED-V} 
& \textbf{Avg.} \\
\midrule

w/o channel axis ($C$) 
& $59.7\pm1.6$ 
& $85.3\pm1.5$ 
& $53.5\pm5.7$ 
& $62.7\pm5.7$ 
& $\underline{62.7\pm9.0}$ 
& $50.6\pm1.3$ 
& $51.8\pm4.4$ 
& $25.0\pm1.3$ 
& $56.40$ \\

w/o short-term axis ($S$) 
& $62.1\pm7.2$ 
& $84.8\pm2.9$ 
& $54.8\pm7.8$ 
& $61.7\pm6.7$ 
& $\mathbf{63.0\pm8.3}$ 
& $52.0\pm4.0$ 
& $53.8\pm2.3$ 
& $27.0\pm1.0$ 
& $57.40$ \\

w/o long-term axis ($L$) 
& $59.6\pm4.7$ 
& $85.3\pm2.5$ 
& $54.9\pm7.3$ 
& $61.8\pm6.3$ 
& $61.7\pm8.7$ 
& $51.3\pm0.8$ 
& $52.7\pm1.2$ 
& $27.0\pm0.6$ 
& $56.78$ \\

Channel axis only ($C$) 
& $55.8\pm11.1$ 
& $84.8\pm2.4$ 
& $52.1\pm9.3$ 
& $61.2\pm8.7$ 
& $60.6\pm6.1$ 
& $51.0\pm2.4$ 
& $51.0\pm5.2$ 
& $27.0\pm3.3$ 
& $55.44$ \\

Short-term axis only ($S$) 
& $50.5\pm2.9$ 
& $85.3\pm1.8$ 
& $54.2\pm7.7$ 
& $63.0\pm11.2$ 
& $59.6\pm7.8$ 
& $50.1\pm2.3$ 
& $52.0\pm0.2$ 
& $25.7\pm1.4$ 
& $55.04$ \\

Long-term axis only ($L$) 
& $48.3\pm1.8$ 
& $84.7\pm1.6$ 
& $53.4\pm6.7$ 
& $59.0\pm3.8$ 
& $62.2\pm7.5$ 
& $50.2\pm1.8$ 
& $51.9\pm0.6$ 
& $26.0\pm3.0$ 
& $54.46$ \\

\midrule

w/o multi-level readout 
& $61.1\pm4.7$ 
& $\mathbf{86.0\pm2.4}$ 
& $55.7\pm6.0$ 
& $59.4\pm5.3$ 
& $62.4\pm7.6$ 
& $49.5\pm2.3$ 
& $\underline{54.4\pm0.6}$ 
& $\underline{29.8\pm5.1}$ 
& $57.30$ \\

w/o cross-axis attention 
& $60.1\pm9.4$ 
& $84.2\pm2.9$ 
& $54.5\pm4.3$ 
& $62.9\pm9.8$ 
& $62.0\pm8.6$ 
& $52.0\pm2.3$ 
& $54.3\pm3.1$ 
& $27.0\pm3.4$ 
& $57.13$ \\

Shared FFN 
& $\underline{62.5\pm6.8}$ 
& $84.8\pm2.6$ 
& $54.6\pm5.9$ 
& $61.8\pm6.9$ 
& $61.0\pm6.4$ 
& $50.9\pm0.8$ 
& $51.6\pm1.8$ 
& $28.8\pm3.9$ 
& $57.01$ \\

Independent attention 
& $57.9\pm7.2$ 
& $85.4\pm2.1$ 
& $\underline{55.8\pm5.1}$ 
& $\underline{63.3\pm8.0}$ 
& $61.7\pm7.7$ 
& $\underline{52.4\pm1.6}$ 
& $\mathbf{56.7\pm1.3}$ 
& $28.4\pm6.1$ 
& $\underline{57.69}$ \\

Sequential $C\rightarrow S\rightarrow L$ 
& $58.3\pm3.6$ 
& $\underline{85.9\pm2.0}$ 
& $52.8\pm7.8$ 
& $\mathbf{64.0\pm5.8}$ 
& $60.5\pm10.0$ 
& $52.0\pm0.9$ 
& $52.4\pm1.2$ 
& $\mathbf{31.1\pm5.6}$ 
& $57.11$ \\

\midrule

\textbf{Full TriDimEEG} 
& $\mathbf{63.8\pm6.1}$ 
& $85.6\pm2.2$ 
& $\mathbf{56.6\pm2.7}$ 
& $60.2\pm7.2$ 
& $60.4\pm8.0$ 
& $\mathbf{53.0\pm2.9}$ 
& $52.9\pm2.2$ 
& $29.0\pm4.3$ 
& $\mathbf{57.70}$ \\

\bottomrule
\end{tabular*}
\par
\endgroup
\caption{Ablation accuracy (\%, mean $\pm$ standard deviation). 
BCI-2A and PNet-MI denote BCI-IV-2A and PhysioNet-MI. 
The best and second-best results in each column are highlighted in bold and 
underlined, respectively. Avg. averages all eight datasets.}
\label{tab:ablation}
\end{table*}

For model design choices, using only the final layer lowers average accuracy
by $0.7\%$ relative to the Full model. Removing cross-axis attention and
sharing one FFN across axes cause relative reductions of $1.0\%$ and
$1.2\%$, respectively, supporting the advantage of combining cross-axis interaction and axis-specific
transformation. Independent attention reaches a comparable performance of $57.69\%$ but uses twice as
many parameters in attention modules, showing that parameter sharing preserves accuracy efficiently.
Sequential $C\rightarrow S\rightarrow L$ processing lowers average accuracy
by $1.0\%$ relative to the Full model,
supporting the advantage of parallel integration over sequential processing.

\subsection{Interpretation analysis}

To examine whether the three TriDim axes capture physiologically meaningful
and class-specific EEG representations, we conduct perturbation analyses along the channel, within-patch time, and across-patch time dimensions. These
three perturbations characterize the spatial, local spectral, and temporal
information used by TriDimEEG for motor imagery decoding, respectively.

\begin{figure}[!t]
  \centering
  \IfFileExists{Figures/jieshi.jpg}{%
    \includegraphics[width=0.90\linewidth]
    {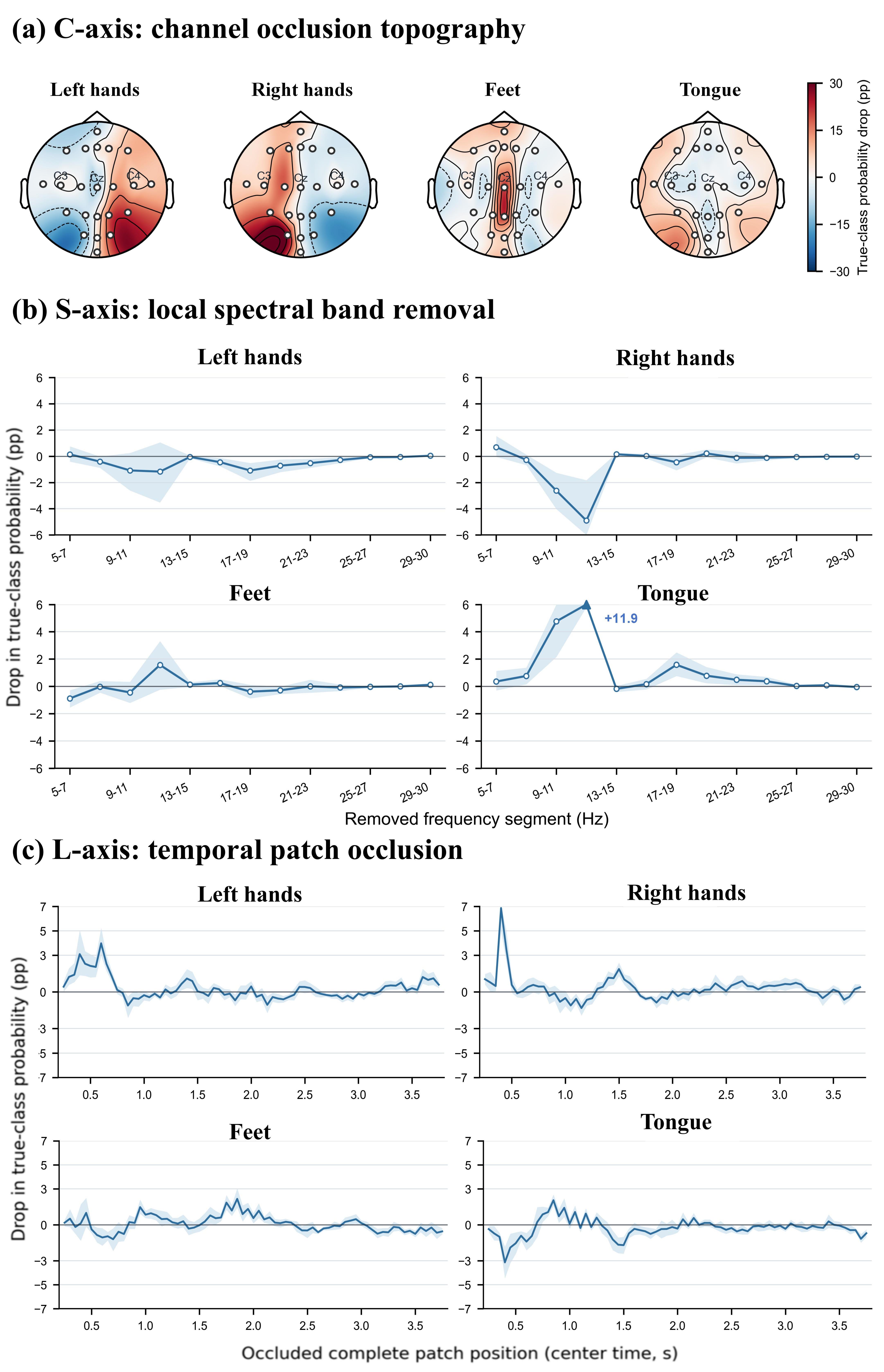}%
  }{%
    \fbox{%
      \parbox[c][0.32\textheight][c]{0.94\linewidth}
    }%
  }
  \caption{Class-conditional C/S/L-axis perturbations on BCI-IV-2A using channel, 2-Hz spectral, and 500-ms patch occlusion. Shading denotes 95\% confidence intervals across nine held-out subjects.}
  \label{fig:three_axis_interp}
\end{figure}

We analyze class-conditional sensitivity on the patched representation
$X^{\mathrm{patch}}\in\mathbb{R}^{B\times C_{in}\times S_{in}\times L_{in}}$. Using 500-ms patches with a 50-ms stride, channel
perturbation zeros one electrode across all patch positions, spectral
perturbation removes one 2-Hz component within every local patch, and
temporal perturbation zeros one complete patch position. We report the signed drop
in true-class probability over nine leave-one-subject-out models, with
95\% subject-bootstrap confidence intervals in
Figure~\ref{fig:three_axis_interp}.

\textbf{Channel occlusion} produces class-dependent scalp patterns
(Figure~\ref{fig:three_axis_interp}(a)). The hand classes predominantly rely
on contralateral centroparietal regions, whereas the feet class shows the
strongest dependence near the central midline. \textbf{Spectral perturbation} shows its largest class-specific effects around
11--13~Hz (Figure~\ref{fig:three_axis_interp}(b)), overlapping the
sensorimotor $\mu$ rhythm. The effect is strongly positive for tongue,
moderately positive for feet, and negative for right hand, indicating that
the same spectral component provides different evidence across classes.
After aggregating the signed drops across classes within each subject, the
overall response retains a clear 11--13~Hz peak
($1.66$~pp; 95\% CI: $0.44$--$3.14$~pp), while other frequencies remain
close to zero. \textbf{Temporal patch occlusion} is strongest during the early trial period for the
hand classes (Figure~\ref{fig:three_axis_interp}(c)). Feet exhibits a later peak around 1.8--2.0~s,
whereas tongue shows an early negative response followed by weaker positive
effects. Together, these results show that TriDimEEG can capture task-dependent spatial, local spectral, and temporal representations.

\section{Conclusion}

We introduced TriDim, a shape-preserving tri-axis block that jointly models
channel organization, short-term waveform structure, and long-term temporal
context through progressive axis-specific refinement and cross-axis
interaction. Under strict cross-subject evaluation across eight EEG
datasets, TriDimEEG achieved the highest average accuracy and best average
rank among 15 models, with ablation and perturbation analyses confirming the
complementary contributions of the three axes. As a drop-in replacement for
Transformer blocks in three pretrained EEG encoders, TriDim achieved a mean
relative improvement of $7.4\%$ in average downstream accuracy over the
corresponding original encoders and reduced parameter
counts. These results demonstrate that TriDim is an effective and reusable
backbone block for task-specific and pretrained EEG models.

\section{Acknowledgments}

This work was supported by the National Natural Science Foundation of
China (62606214, 3254100307, and 62472206), the National Science and
Technology Major Project (2021ZD0200500), the National Key R\&D Program
of China (2025YFC3410000), the Guangdong Basic and Applied Basic Research
Foundation (2026A1515010121), the Guangdong S\&T Program
(2026B0101110003), the Shenzhen Science and Technology Innovation
Committee (RCYX20231211090405003 and JCYJ20220818100213029), the
Guangdong Provincial Key Laboratory of Advanced Biomaterials
(2022B1212010003), and the open research fund of the Guangdong Provincial
Key Laboratory of Mathematical and Neural Dynamical Systems and the
Center for Computational Science and Engineering at Southern University
of Science and Technology.

\FloatBarrier
\bibliographystyle{abbrvnat}
\bibliography{references}

\flushcolsend
\onecolumn
\input{TriDim_Supplementary_Material}

\end{document}

%% file: TriDim_Supplementary_Material.tex
\begingroup
\appendix
\setcounter{secnumdepth}{1}
\setcounter{table}{0}
\setcounter{figure}{0}
\setcounter{equation}{0}
\renewcommand{\thetable}{\arabic{table}}
\renewcommand{\thefigure}{S\arabic{figure}}
\renewcommand{\theequation}{S\arabic{equation}}
\renewcommand{\theHtable}{supp.\arabic{table}}
\renewcommand{\theHfigure}{supp.\arabic{figure}}
\renewcommand{\theHequation}{supp.\arabic{equation}}
\setlength{\parindent}{1em}
\setlength{\parskip}{0pt}
\setlength{\textfloatsep}{10pt plus 2pt minus 2pt}
\setlength{\floatsep}{8pt plus 2pt minus 2pt}
\setlength{\intextsep}{10pt plus 2pt minus 2pt}
\renewcommand{\arraystretch}{1.08}
\titleformat{\section}{\normalfont\fontsize{11}{13}\selectfont\bfseries\centering}{\thesection.}{0.5em}{}
\titlespacing*{\section}{0pt}{12pt plus 2pt minus 2pt}{6pt}
\titleformat{\paragraph}[runin]{\normalfont\normalsize\bfseries}{}{0pt}{}
\titlespacing*{\paragraph}{0pt}{8pt plus 2pt minus 2pt}{0.5em}
\captionsetup{font=normalsize,labelfont=bf,justification=justified,singlelinecheck=false,skip=4pt}
\captionsetup[table]{name=Supplementary Table}
\begin{center}
{\fontsize{14}{17}\selectfont\bfseries Supplementary Material\par}
\end{center}

\section{Dataset and Preprocessing Details}
\label{sec:data}

Table~\ref{tab:dataset_statistics} of the main text summarizes the eight processed datasets.
All datasets are publicly available for research purposes and are cited in
the main text. Raw recordings are preprocessed with a unified pipeline:
resampling to 200~Hz, a 0.3--75~Hz band-pass filter, and a notch filter at
the local power-line frequency. AD65, FACED, SEED, and SEED-V are segmented
into 10-s windows; BCI-IV-2A, SHU-MI, and PhysioNet-MI use 4-s windows;
SleepEDF uses standard 30-s sleep epochs. Per-sample, per-channel temporal
normalization is performed by the InstanceTimeNorm layer at the model input,
so no dataset-level statistics are computed or leaked across subjects.

For SEED-V we use the official 10-s benchmark segments; the label of each
10-s segment is inherited from its source 1-s annotation block. For
SleepEDF we follow the supplied session-level protocol (153 sessions split
122/15/16) and retain at most 40 30-s epochs per session--sleep-stage pair
in the training subset only (330{,}374 $\rightarrow$ 22{,}861 training
epochs; the frozen manifests record the exact retained indices), which both
bounds the training-set size and balances the five sleep stages
(4{,}880/4{,}787/3{,}450/4{,}880/4{,}864 epochs for seed~5). Validation and
test sessions are kept complete without truncation or downsampling
(41{,}150 and 43{,}437 epochs), preserving the natural class distribution.
The same retained-epoch manifests are used by every compared method on
SleepEDF.

\section{Evaluation Protocol}
\label{sec:protocol}

All methods are evaluated under a strict subject-independent protocol.
Subjects are partitioned into training, validation, and test sets with an
approximate 8:1:1 ratio, stratified by class where applicable; every
recording, session, and segment of a subject is assigned to exactly one
subset. Subject-level split manifests are generated before training, frozen
as immutable JSON files, and shared by all compared methods, so every model
is trained and tested on identical subject partitions. The frozen manifests
for SEED-V and SleepEDF are shipped with the released code, together with
their SHA-256 checksums; the remaining datasets use a seeded stratified
subject splitter (train ratio 0.8, validation ratio 0.1) whose outputs are
equally immutable for a given seed.

Each configuration is run under three independent subject-level splits
generated by random seeds 5, 42, and 43. Within each run, the checkpoint
with the highest validation accuracy is selected for final testing. We
report the mean and sample standard deviation (ddof = 1) of test accuracy
across the three splits. As aggregate measures across the eight datasets we
report (i)~average accuracy and (ii)~average rank, where each model is
ranked per dataset by mean accuracy (rank 1 = best) and the ranks are
averaged; a lower average rank indicates better overall performance.
Macro-F1 scores of the Full model and all ablation variants are
reported in Supplementary Table~\ref{tab:persplit}; per-run records of every metric
are included in the released result files.

Performance changes in the text are expressed relative to the stated
reference: $100(A_{\mathrm{new}}-A_{\mathrm{ref}})/A_{\mathrm{ref}}\%$,
where $A$ is the accuracy or macro-F1 for the comparison. For cross-dataset
comparisons, $A$ is the mean across the eight datasets. Ablation reductions
use the Full model as the reference. Paired statistical differences and
perturbation-induced probability changes retain their percentage-point units.

\section{Macro-F1 Result Records}
\label{app:persplit}

Supplementary Table~\ref{tab:persplit} reports the macro-F1 score of the Full model
and all eleven ablation variants on each of the eight datasets, as the
mean $\pm$ sample standard deviation across the three subject-level
splits (seeds 5/42/43); the underlying runs are exactly those behind
Tables~\ref{tab:main_results} and~\ref{tab:ablation} of the main text. Per-split accuracy and macro-F1
records of every run, including the compared baselines, are included
in the released result files. Average macro-F1 and average rank are computed
across all eight datasets; a lower average rank indicates better overall
cross-dataset ranking.

\begin{table}[htbp]
\centering
\begingroup
\footnotesize
\caption{Macro-F1 (\%) of Full TriDimEEG and all ablation variants on the
eight datasets. Each cell reports the mean $\pm$ sample standard
deviation (ddof = 1) across the three subject-level splits
(seeds 5/42/43); the underlying runs are exactly those behind
Tables~\ref{tab:main_results} and~\ref{tab:ablation} of the main text.
The best and second-best results in each column are highlighted in bold and
underlined, respectively.}
\label{tab:persplit}
\setlength{\tabcolsep}{0.4pt}
\setlength{\medmuskip}{1mu}
\renewcommand{\arraystretch}{1.05}
\begin{tabular*}{\textwidth}{@{\extracolsep{\fill}}lccccccccc@{}}
\toprule
\textbf{Variant} & \textbf{AD65} & \textbf{SleepEDF} & \textbf{BCI-2A}
& \textbf{SHU-MI} & \textbf{PNet-MI} & \textbf{FACED} & \textbf{SEED}
& \textbf{SEED-V} & \textbf{Avg. $\uparrow$} \\
\midrule
w/o channel axis ($C$)
& $51.2\pm8.05$ & $65.9\pm3.26$ & $53.1\pm5.89$ & $62.2\pm5.85$
& $\underline{62.9\pm8.93}$ & $51.0\pm1.34$ & $51.0\pm5.20$ & $18.0\pm8.49$
& $51.91$ \\
w/o short-term axis ($S$)
& $55.3\pm13.8$ & $65.9\pm5.08$ & $54.6\pm7.58$ & $61.6\pm6.67$
& $\mathbf{63.0\pm8.55}$ & $52.5\pm3.97$ & $52.8\pm3.66$ & $20.2\pm10.3$
& $53.24$ \\
w/o long-term axis ($L$)
& $54.8\pm8.50$ & $66.4\pm4.93$ & $54.5\pm7.67$ & $61.7\pm6.26$
& $61.7\pm8.45$ & $51.7\pm0.810$ & $51.9\pm1.42$ & $20.4\pm10.5$
& $52.89$ \\
Channel axis only ($C$)
& $44.8\pm15.4$ & $66.7\pm3.99$ & $51.7\pm9.48$ & $61.1\pm8.70$
& $60.6\pm6.41$ & $51.2\pm2.60$ & $49.8\pm6.75$ & $23.6\pm2.38$
& $51.19$ \\
Short-term axis only ($S$)
& $43.0\pm4.61$ & $66.1\pm3.37$ & $53.9\pm7.85$ & $62.6\pm11.5$
& $59.3\pm8.12$ & $50.4\pm2.21$ & $50.5\pm2.50$ & $17.9\pm8.39$
& $50.46$ \\
Long-term axis only ($L$)
& $40.8\pm4.40$ & $65.4\pm3.39$ & $53.0\pm7.19$ & $58.7\pm4.04$
& $62.5\pm7.50$ & $50.6\pm1.63$ & $51.3\pm1.10$ & $24.6\pm2.63$
& $50.86$ \\
\midrule
w/o multi-level readout
& $54.9\pm10.2$ & $\mathbf{67.9\pm3.80}$ & $\underline{55.5\pm5.97}$ & $59.3\pm5.39$
& $62.3\pm7.71$ & $49.8\pm2.33$ & $\underline{53.8\pm0.950}$ & $28.0\pm5.13$
& $53.94$ \\
w/o cross-axis attention
& $\underline{56.7\pm12.1}$ & $65.1\pm5.33$ & $54.0\pm4.25$ & $62.6\pm10.1$
& $62.1\pm8.67$ & $52.3\pm2.22$ & $53.4\pm3.33$ & $26.9\pm3.41$
& $\underline{54.14}$ \\
Shared FFN
& $56.2\pm13.5$ & $65.4\pm4.50$ & $53.7\pm5.68$ & $61.1\pm7.53$
& $61.2\pm6.15$ & $51.4\pm0.740$ & $51.2\pm1.19$ & $26.6\pm3.54$
& $53.35$ \\
Independent attention
& $49.2\pm3.45$ & $66.4\pm4.02$ & $\underline{55.5\pm4.93}$ & $\underline{63.2\pm7.89}$
& $61.7\pm7.82$ & $\underline{52.7\pm1.74}$ & $\mathbf{56.0\pm1.56}$ & $26.5\pm7.21$
& $53.90$ \\
Sequential $C{\rightarrow}S{\rightarrow}L$
& $51.6\pm5.54$ & $\underline{67.5\pm3.01}$ & $52.0\pm8.16$ & $\mathbf{63.9\pm5.87}$
& $60.6\pm10.0$ & $52.3\pm0.950$ & $51.6\pm0.190$ & $\mathbf{30.8\pm6.24}$
& $53.79$ \\
\midrule
\textbf{Full TriDimEEG}
& $\mathbf{60.4\pm7.62}$ & $66.9\pm3.17$ & $\mathbf{56.1\pm3.11}$ & $59.5\pm7.16$
& $60.6\pm7.83$ & $\mathbf{53.3\pm3.06}$ & $51.2\pm1.95$ & $\underline{28.2\pm4.54}$
& $\mathbf{54.53}$ \\
\bottomrule
\end{tabular*}
\endgroup
\end{table}

\section{Leave-One-Subject-Out Evaluation}
\label{app:loso}

The 8:1:1 subject partition used in the main table is retained for
consistency across all eight datasets, but on small-cohort datasets it
yields test sets of one to three subjects. We therefore additionally
evaluate the Full TriDimEEG model under leave-one-subject-out (LOSO)
cross-validation on the four small-cohort datasets: BCI-IV-2A (9
subjects), SHU-MI (25 subjects), SEED (15 subjects), and SEED-V (16
subjects). BCI-IV-2A is the most affected case and is also the dataset
with the most established LOSO convention; we report its complete
fold$\times$seed grid below and summarize the remaining three datasets
at the fold level.

\paragraph{Protocol.} For fold $k \in \{1,\dots,9\}$, subject A0$k$ is the
test subject, subject A0$((k \bmod 9)+1)$ is the validation subject, and the
remaining seven subjects form the training set. Every fold is trained under
the same three seeds (5, 42, 43) and the identical per-dataset
configuration as the main-table BCI-IV-2A runs (Supplementary
Tables~\ref{tab:arch} and~\ref{tab:train}), with validation-based checkpoint selection. The
fold manifests are frozen JSON files with disjoint train/validation/test
trial indices, shipped with the released code. The LOSO study uses the
fixed BCI-IV-2A trial asset of 5{,}088 trials (subject A04 contributes 480
trials; all other subjects 576); Table~\ref{tab:dataset_statistics} of the main text reports the raw
segment count (5{,}184 = $9 \times 576$), and the released trial asset
excludes 96 trials of subject A04, yielding the fixed 5{,}088-trial set
used by all BCI-IV-2A experiments. The perturbation analysis uses the same nine
seed-42 LOSO subject partitions, but retrains a separate set of models
with 500-ms patches and a 50-ms stride. These interpretation models are
therefore distinct from the 200-ms-patch models reported in Supplementary
Table~\ref{tab:loso}.

\begin{table}[htbp]
\centering
\caption{LOSO cross-subject test accuracy (\%) of Full TriDimEEG on
BCI-IV-2A. Each cell is one trained model; mean $\pm$ std is computed
across the three seeds per fold. The overall row reports the mean $\pm$
std across all 27 fold$\times$seed runs.}
\label{tab:loso}
\setlength{\tabcolsep}{4pt}
\begin{tabular*}{\textwidth}{@{\extracolsep{\fill}}lccccc@{}}
\toprule
Fold (test subj.) & seed 5 & seed 42 & seed 43 & mean $\pm$ std \\
\midrule
fold 1 (A01) & 63.7 & 38.7 & 47.1 & $49.8 \pm 12.7$ \\
fold 2 (A02) & 45.8 & 39.6 & 41.8 & $42.4 \pm 3.16$ \\
fold 3 (A03) & 65.3 & 57.6 & 61.5 & $61.5 \pm 3.82$ \\
fold 4 (A04) & 45.4 & 41.7 & 43.3 & $43.5 \pm 1.88$ \\
fold 5 (A05) & 51.6 & 58.2 & 51.0 & $53.6 \pm 3.97$ \\
fold 6 (A06) & 48.3 & 45.5 & 47.7 & $47.2 \pm 1.48$ \\
fold 7 (A07) & 57.1 & 55.9 & 61.1 & $58.0 \pm 2.72$ \\
fold 8 (A08) & 56.8 & 50.9 & 52.1 & $53.2 \pm 3.12$ \\
fold 9 (A09) & 55.7 & 45.1 & 52.1 & $51.0 \pm 5.38$ \\
\midrule
Overall & \multicolumn{3}{c}{per-seed fold means: 54.4 / 48.1 / 50.9}
        & $51.1 \pm 7.49$ \\
\bottomrule
\end{tabular*}
\end{table}

\paragraph{Results.} Supplementary Table~\ref{tab:loso} reports the complete
fold$\times$seed grid. TriDim obtains $51.1 \pm 7.49$\% under LOSO, below
its 8:1:1 result ($56.6 \pm 2.71$\%); this gap is expected because LOSO
evaluates every subject, including the hardest ones, whereas the 8:1:1
protocol tests on a single drawn subject per split. Per-fold means range
from $42.4$\% (A02) to $61.5$\% (A03), reflecting the well-documented
subject difficulty spread of this benchmark. We report this LOSO study as a
protocol-robustness check; the main-table protocol is unchanged for
cross-dataset comparability.

\paragraph{LOSO on SHU-MI, SEED, and SEED-V.}
We additionally ran LOSO cross-validation of the Full model (seed 42,
identical per-dataset configurations as the main table) on the three
remaining small-cohort datasets, with every subject serving once as the
test subject and no subject leakage between training and testing.
Supplementary Table~\ref{tab:loso_extra} summarizes the fold-level results. On all
three datasets the LOSO accuracy is consistent with the 8:1:1 three-split
results of the main table ($60.2 \pm 7.16$ on SHU-MI, $52.9 \pm 2.22$
on SEED, and $29.0 \pm 4.34$ on SEED-V), indicating that the main
conclusions are robust to the choice of subject-partition protocol.

\begin{table}[htbp]
\centering
\caption{LOSO cross-subject results of Full TriDimEEG (seed 42) on SHU-MI,
SEED, and SEED-V. Accuracy and macro-F1 (\%) are reported as mean $\pm$
std across folds (one test subject per fold). The 3-split column recalls
the main-table accuracy for reference.}
\label{tab:loso_extra}
\setlength{\tabcolsep}{3.5pt}
\begin{tabular*}{\textwidth}{@{\extracolsep{\fill}}lccc@{}}
\toprule
Dataset & LOSO acc. & LOSO macro-F1 & 3-split acc. \\
\midrule
SHU-MI & $60.8 \pm 12.4$ & $59.9 \pm 12.8$ & $60.2 \pm 7.16$ \\
SEED   & $54.5 \pm 7.93$  & $51.8 \pm 9.44$  & $52.9 \pm 2.22$ \\
SEED-V & $29.3 \pm 4.86$  & $26.9 \pm 6.87$  & $29.0 \pm 4.34$ \\
\bottomrule
\end{tabular*}
\end{table}

\section{Baseline Compatibility and Adaptation Details}
\label{app:baselines}

All baselines are evaluated under the same frozen subject-level manifests,
the same validation-based checkpoint selection, and the same early-stopping
protocol as TriDim. Input-shape adaptation follows each official
implementation without altering the subject partition: recordings are
resampled, filtered, and segmented by the unified pipeline
(see the \emph{Dataset and Preprocessing Details} section), and each baseline receives its inputs in the channel
layout its official code defines. When a dataset montage cannot supply the
channels or derivations a baseline requires, the corresponding inputs are
zero-filled; no method-specific subject re-partitioning, re-segmentation,
or label remapping is applied.

\section{Pretrained-Encoder Integration: Protocol and Per-Split Results}
\label{app:plugin}

\paragraph{Integration protocol.} For each host encoder (REVE, CBraMod,
CSBrain), the original backbone blocks are replaced with TriDim blocks and
the modified encoder is \emph{re-pretrained from scratch} with the host's
original pretraining objective (masked reconstruction). For a controlled
comparison, both the original and the TriDim-based backbones are pretrained
on the same $2{,}000$ samples selected from the TUH corpus (TUH-2K), with
identical pretraining schedules and otherwise unchanged components (input
frontend, pretraining head, positional schemes). Downstream evaluation then
loads the re-pretrained weights and fine-tunes \emph{all} parameters
end-to-end with a newly initialized linear classification head---no module
is frozen, and shape-compatible host tensors are loaded with a non-strict
state-dict mapping. Because this reduced TUH-2K pretraining set differs
from the original released checkpoints used in
Table~\ref{tab:main_results} of the main text, the absolute numbers of
this section are not directly comparable with the main table; the
comparison of interest is original versus TriDim-based backbone within each
encoder under identical settings.

\paragraph{Tensor interface.} TriDim operates on a $[B, C, S, L]$ tensor;
each host supplies its own grid mapping. For CBraMod, the backbone is
natively four-dimensional $[B, C, P, d_{\mathrm{model}}]$ with
$d_{\mathrm{model}} = 200$; the channel, patch, and $d_{\mathrm{model}}$
axes map directly onto the $C$, $S$, and $L$ axes of TriDim, and the
convolutional stem, FFT features, and ACPE positional encoding are
retained. For CSBrain, the backbone tensor $[B, D, C, P]$ is mapped with
the feature dimension $D$ as the channel axis, the brain-region axis as
$S$, and the patch axis as $L$. For REVE, the flattened token sequence is
rearranged into a two-dimensional token grid
(\texttt{b (c h) d $\rightarrow$ b c h d}), with the host's Fourier
3D-coordinate and time-index positional embeddings retained on the tokens.
In all three cases no information is discarded by the mapping; the token
count and feature width of the host are preserved.

\paragraph{Result records.} Supplementary Table~\ref{tab:plugin} lists the per-dataset
mean test accuracy of the original and TriDim-based encoders under the
TUH-2K setting. Per-split values are included in the released result files.

\begin{table}[htbp]
\centering
\caption{Per-dataset mean test accuracy (\%) of original and TriDim-based
encoders under the TUH-2K pretraining setting, averaged over the three
subject-level splits (seeds 5/42/43).}
\label{tab:plugin}
\setlength{\tabcolsep}{7pt}
\begin{tabular*}{\textwidth}{@{\extracolsep{\fill}}lcccccc@{}}
\toprule
& \multicolumn{2}{c}{REVE} & \multicolumn{2}{c}{CBraMod} & \multicolumn{2}{c}{CSBrain} \\
\cmidrule(lr){2-3}\cmidrule(lr){4-5}\cmidrule(lr){6-7}
Dataset & orig. & +TriDim & orig. & +TriDim & orig. & +TriDim \\
\midrule
AD65         & 51.4 & 55.4 & 53.0 & 59.1 & 51.4 & 55.9 \\
SleepEDF     & 70.2 & 84.4 & 84.2 & 83.5 & 81.6 & 86.0 \\
BCI-IV-2A    & 37.3 & 45.6 & 44.5 & 44.9 & 41.2 & 46.6 \\
SHU-MI       & 59.1 & 61.2 & 57.6 & 60.3 & 59.1 & 63.5 \\
PhysioNet-MI & 57.5 & 57.8 & 57.6 & 59.4 & 56.5 & 59.0 \\
FACED        & 43.2 & 48.5 & 65.5 & 57.8 & 65.6 & 63.7 \\
SEED         & 40.3 & 38.6 & 37.8 & 41.6 & 39.0 & 42.7 \\
SEED-V       & 16.7 & 29.1 & 21.4 & 27.8 & 21.9 & 29.3 \\
\midrule
Average      & 47.0 & 52.6 & 52.7 & 54.3 & 52.0 & 55.8 \\
\bottomrule
\end{tabular*}
\end{table}

Under this controlled evaluation, TriDim-based re-pretraining yields relative
improvements in average downstream accuracy of $11.91\%$, $3.05\%$, and
$7.35\%$ over the original REVE, CBraMod, and CSBrain encoders, respectively.
For each encoder, the relative improvement is computed from its mean
accuracy across the eight datasets; averaging the three encoder-level
relative improvements gives $7.4\%$. This is not an average of per-dataset
relative improvements. Parameter counts are reduced by $32.0\%$, $17.0\%$, and
$47.3\%$---matching the values reported in the main text. Gains are
consistent across task categories for REVE and CSBrain; the CBraMod variant
trades small decreases on FACED and SleepEDF for gains on the remaining six
datasets.

\section{Complete Hyperparameter Configurations}
\label{sec:hyper}

Supplementary Tables~\ref{tab:arch} and~\ref{tab:train} list the exact final
hyperparameters used for the Full TriDimEEG model on every dataset, taken
verbatim from the released per-dataset YAML configuration files.
Dataset-specific values were selected exclusively according to validation
accuracy during development; all ablation variants reuse the same
per-dataset configuration with only the structural change described in
the following section. The multi-level tri-axis readout is enabled in all
main experiments.

\begin{table}[htbp]
\centering
\caption{Architecture hyperparameters of Full TriDimEEG per dataset.
$C/S/L$: latent axis sizes (basis dims); $D$: patch embedding width;
$N$: number of TriDim blocks; $p_a$: maximum DropPath rates of the
attention views and the FFN branch ($p_c/p_s/p_l/p_{\mathrm{ffn}}$).
The last column states how the DropPath rate is applied across the
$N$ blocks: \emph{constant} holds the rate at $p_a$ in every block,
while \emph{linear $0{\to}p_a$} ramps it linearly from $0$ to $p_a$
across blocks.}
\label{tab:arch}
\setlength{\tabcolsep}{3pt}
\begin{tabular*}{\textwidth}{@{\extracolsep{\fill}}lcccccccc@{}}
\toprule
Dataset & Channels & Patch len/stride & $C$ & $S$ & $L$ & $D$ & Heads & $N$ \\
\midrule
AD65          & 19 & 80/20 & 96 & 32 & 32 & 128 & 4 & 4 \\
SleepEDF      &  2 & 80/20 & 32 & 16 & 16 & 128 & 4 & 4 \\
BCI-IV-2A     & 22 & 40/10 & 96 & 32 & 32 & 128 & 4 & 3 \\
SHU-MI        & 32 & 32/8  & 48 & 16 & 16 & 128 & 4 & 3 \\
PhysioNet-MI  & 64 & 80/20 & 96 & 32 & 32 & 256 & 4 & 6 \\
FACED         & 32 & 80/20 & 96 & 32 & 32 & 256 & 4 & 6 \\
SEED          & 62 & 80/80 & 96 & 32 & 32 & 128 & 4 & 3 \\
SEED-V        & 62 & 64/16 & 96 & 32 & 32 & 256 & 4 & 6 \\
\bottomrule
\end{tabular*}

\medskip
\begin{tabular*}{\textwidth}{@{\extracolsep{\fill}}lccc@{}}
\toprule
Dataset & Dropout & $p_c/p_s/p_l/p_{\mathrm{ffn}}$ & DropPath schedule \\
\midrule
AD65          & 0.20 & 0.10/0.02/0.08/0.02 & constant \\
SleepEDF      & 0.15 & 0.25/0.05/0.15/0.05 & linear $0{\to}p_a$ \\
BCI-IV-2A     & 0.10 & 0.10/0.02/0.08/0.02 & constant \\
SHU-MI        & 0.20 & 0.10/0.02/0.08/0.02 & constant \\
PhysioNet-MI  & 0.30 & 0.25/0.05/0.15/0.05 & linear $0{\to}p_a$ \\
FACED         & 0.15 & 0.10/0.02/0.08/0.02 & constant \\
SEED          & 0.15 & 0.10/0.02/0.08/0.02 & constant \\
SEED-V        & 0.20 & 0.10/0.02/0.08/0.02 & constant \\
\bottomrule
\end{tabular*}
\end{table}

\begin{table}[htbp]
\centering
\caption{Training hyperparameters of Full TriDimEEG per dataset.
All runs use AdamW, cosine learning-rate decay with a 5-epoch linear
warm-up, gradient clipping at norm 1.0, early stopping on validation
accuracy, and checkpoint selection by validation accuracy.}
\label{tab:train}
\setlength{\tabcolsep}{8pt}
\begin{tabular*}{\textwidth}{@{\extracolsep{\fill}}lccccc@{}}
\toprule
Dataset & Batch size & Learning rate & Weight decay & Max epochs & Patience \\
\midrule
AD65          & 32 & 5$\times$10$^{-4}$ & 0.02 & 100 & 20 \\
SleepEDF      & 32 & 5$\times$10$^{-4}$ & 0.05 & 50  & 50 \\
BCI-IV-2A     & 16 & 5$\times$10$^{-4}$ & 0.01 & 100 & 20 \\
SHU-MI        & 64 & 1$\times$10$^{-4}$ & 0.02 & 100 & 15 \\
PhysioNet-MI  & 32 & 5$\times$10$^{-4}$ & 0.02 & 100 & 20 \\
FACED         & 32 & 5$\times$10$^{-4}$ & 0.00 & 100 & 20 \\
SEED          & 32 & 5$\times$10$^{-4}$ & 0.01 & 100 & 20 \\
SEED-V        & 32 & 5$\times$10$^{-4}$ & 0.05 & 50  & 50 \\
\bottomrule
\end{tabular*}
\end{table}

\section{Ablation Variant Definitions}
\label{sec:ablation}

All ablation variants share the Full model's frontend (InstanceTimeNorm,
patch unfolding, three basis projections), the multi-level tri-axis
readout, the classifier head, and every training hyperparameter; only the
indicated structural element differs. Parameter counts below are measured
under the SEED configuration (Full = 3{,}251{,}600); exact counts vary
slightly with per-dataset configurations.

\begin{itemize}
\item \textbf{w/o channel axis ($C$) / w/o short-term axis ($S$) / w/o
long-term axis ($L$).} The indicated axis is removed from the block: its
cross-axis attention branch, its axis-specific FFN, and the associated
normalizations and DropPath are dropped, and the learned softmax view
fusion is renormalized over the two surviving axis views.

\item \textbf{Channel axis only ($C$) / short-term axis only ($S$) /
long-term axis only ($L$).} Only the indicated axis is retained: the
attention branch and the axis-specific FFN of the kept axis are
unchanged, the corresponding branches of the other two axes are removed,
and the view fusion degenerates to the single surviving view.

\item \textbf{w/o multi-level readout.} The multi-level tri-axis readout
is disabled (\texttt{use\_multi\_level\_readout=False}); the per-level
readout projections and the layer-fusion weights are removed, and the
single tri-axis attention-pooling head is applied to the final block
output only.

\item \textbf{w/o cross-axis attention.} The entire cross-axis attention
sublayer is removed (the three shared axis-attention modules, per-axis
attention normalizations, attention view-fusion weights, attention
LayerScale, and the per-branch DropPath), leaving a purely feed-forward
tri-axis mixer: $x \leftarrow x + \gamma_{\mathrm{ffn}} \odot
\mathrm{FFN}(x)$. $-137{,}865$ parameters.

\item \textbf{Shared FFN.} The three axis-specific FFNs are replaced by one
shared two-layer MLP (same expansion ratio 2). Each axis view is
transposed so the target axis becomes the last dimension, right-zero-padded
up to the channel-axis width, passed through the shared MLP, and sliced
back; padding only ever widens the $S$/$L$ views, so no view is truncated.
$-25{,}152$ parameters.

\item \textbf{Independent attention.} The three parameter-shared attention
modules are replaced by six independent modules, one per (embedding axis,
sequence axis) pair; every other component is unchanged. $+137{,}088$
parameters.

\item \textbf{Sequential $C{\rightarrow}S{\rightarrow}L$.} The three axis
views are applied sequentially in the fixed order $C \rightarrow S
\rightarrow L$ instead of in parallel with learned softmax fusion; each
sub-update reads the tensor updated by the previous one. The view-fusion
weights are removed ($-18$ scalar parameters). During training, one of the
three attention sub-updates is randomly skipped per batch with the
configured probability and the surviving two are scaled by $1.5$, keeping
the expected update identical to the parallel design.
\end{itemize}

\section{Statistical Analysis}
\label{app:stats}

\paragraph{Variation across splits.}
Every entry in Tables~\ref{tab:main_results} and~\ref{tab:ablation} of the main text is the mean $\pm$ sample
standard deviation over the three independent subject-level splits
(seeds 5/42/43), rather than a single run; the corresponding macro-F1
scores are listed in Supplementary Table~\ref{tab:persplit}, and the underlying
per-split values are included in the released result files.

\paragraph{Cross-model omnibus test.}
A Friedman test across the 15 compared models and the eight datasets
(mean accuracy per dataset as the observation unit) rejects the null
hypothesis of equal model performance ($\chi^2(14) = 55.84$, $p < 0.001$),
confirming that the per-dataset ranking differences are systematic rather
than random.

\paragraph{Cross-dataset pairwise tests.}
We compare TriDim with the strongest baselines using a two-sided Wilcoxon
signed-rank test paired by dataset ($n = 8$, exact).
TriDim wins 5/8 datasets against the second-best model EEGDeformer
($W = 11$, $p = 0.383$), 6/8 against REVE ($W = 6$, $p = 0.109$), and 7/8
against CBraMod ($W = 5$, $p = 0.078$). With only eight paired datasets the
test has limited power (the smallest attainable two-sided $p$ is $0.0078$,
and only for an 8/8 win), so we interpret the aggregate performance
improvements ($4.3\%$ relative improvement in average accuracy over EEGDeformer, together
with the better average rank) as the primary evidence, and report the test
outcomes for completeness rather than as the basis of our claims.

\paragraph{Split-level tests for ablations.}
For the ablation study we exploit the fact that every variant is evaluated
on the \emph{same} frozen split manifests as the Full model: we pair
per-split test accuracies by (dataset, split), giving $n = 8 \times 3 = 24$
paired observations per variant, and apply a two-sided Wilcoxon signed-rank
test (normal approximation with continuity correction).
Supplementary Table~\ref{tab:ablation_stats} reports the mean paired difference
(Full $-$ variant) and the test outcome. Removing any single axis does not
significantly degrade performance ($p = 0.089$ for $C$, a marginal trend;
$p = 0.70$ for $S$; $p = 0.62$ for $L$), whereas restricting the model to a
single axis does: the $S$-only and $L$-only variants are significantly
worse than Full ($p = 0.012$ and $p = 0.015$), with $C$-only a marginal
trend ($p = 0.10$). The structural variants (multi-level readout, cross-axis
attention, shared FFN, sequential processing) each reduce average accuracy
by approximately $0.7$--$1.2\%$ relative to Full without reaching
significance, and independent (unshared) attention performs essentially
on par with Full, indicating that
parameter sharing is performance-neutral. We therefore describe the
tri-axis design as improving cross-task robustness---the Full model is the
most consistent variant across datasets rather than the per-dataset
winner---and avoid claiming per-variant significant differences where the
paired test does not support them.

\begin{table}[htbp]
\centering
\caption{Split-level paired Wilcoxon signed-rank tests, Full TriDimEEG versus
each ablation variant, paired by (dataset, split) with $n = 24$. Mean
$\Delta$ is the mean paired accuracy difference (Full $-$ variant) in
percentage points; positive values favor Full. Mean $\Delta$ is computed
from unrounded per-split values and may therefore differ by up to
$0.01$~pp from the difference of the rounded table entries.}
\label{tab:ablation_stats}
\begin{tabularx}{\textwidth}{@{}X>{\centering\arraybackslash}p{0.22\textwidth}>{\centering\arraybackslash}p{0.16\textwidth}@{}}
\toprule
Variant & Mean $\Delta$ (pp) & Wilcoxon $p$ \\
\midrule
w/o channel axis ($C$)        & $+1.30$ & $0.089$ \\
w/o short-term axis ($S$) & $+0.31$ & $0.700$ \\
w/o long-term axis ($L$)& $+0.92$ & $0.617$ \\
Channel axis only ($C$)       & $+2.26$ & $0.100$ \\
Short-term axis only ($S$)& $+2.66$ & $0.012$ \\
Long-term axis only ($L$)& $+3.24$ & $0.015$ \\
w/o multi-level readout       & $+0.40$ & $0.786$ \\
w/o cross-axis attention      & $+0.57$ & $0.637$ \\
Shared FFN                    & $+0.69$ & $0.539$ \\
Independent attention         & $+0.00$ & $0.353$ \\
Sequential $C{\rightarrow}S{\rightarrow}L$ & $+0.59$ & $0.867$ \\
\bottomrule
\end{tabularx}
\end{table}

\section{Sensitivity Analysis Methodology}
\label{app:sensitivity}

\paragraph{Perturbation analysis.}
The axis-aligned perturbation study shown in Figure~\ref{fig:three_axis_interp}
uses nine seed-42 LOSO models trained specifically for the interpretation
analysis with 500-ms patches and a 50-ms stride. These models use the same
LOSO subject partitions as the protocol analysis above, but differ from
the 200-ms-patch BCI-IV-2A models used in the main benchmark and
Supplementary Table~\ref{tab:loso}. Each subject is scored by a model that
never saw that subject during training or checkpoint selection.
Perturbations are applied after patchification to the patched
representation rather than directly to the raw EEG signal. Three
perturbation families are evaluated class-conditionally:
(i)~\emph{electrode occlusion}, zeroing one of the 22 channels across all
patch positions; (ii)~\emph{frequency-band occlusion}, removing one 2-Hz
spectral component within each 500-ms local patch; and
(iii)~\emph{temporal patch occlusion}, zeroing one complete 500-ms patch
while leaving the overlapping neighboring patches unchanged. With a
50-ms stride over each 4-s trial, the temporal analysis contains 71 patch
positions. We record the signed drop in true-class probability (in
percentage points) relative to the unperturbed input. For each condition,
the subject-level mean drops are pooled across the nine held-out subjects,
and 95\% confidence intervals are obtained from a 10{,}000-draw bootstrap
over subjects ($n = 9$). The complete per-condition summary tables are
included in the released result files.

\paragraph{Alignment with the main configuration.}
The interpretation analysis deliberately uses a 500-ms patch length with
a 50-ms stride, whereas the main BCI-IV-2A benchmark uses 200-ms patches
with the same 50-ms stride. The 500-ms configuration provides the patch
representation used for all three perturbation analyses in
Figure~\ref{fig:three_axis_interp} and yields 71 temporal occlusion
positions over each 4-s trial.

\paragraph{Shuffled-label controls.}
As a negative control, we retrain the same nine LOSO models with the
\emph{training} labels randomly permuted (fixed permutation, seed 42;
validation and test labels untouched) under the identical configuration.
These models perform at chance on the held-out subjects (mean test accuracy
$26.1\%$ over the nine folds, versus $25\%$ chance), confirming that no
task-relevant structure is learned. Their perturbation profile collapses:
the mean absolute true-class-probability drop shrinks from $5.71$ to
$0.71$~pp for electrode occlusion, from $0.41$ to $0.14$~pp for
frequency-band occlusion, and from $0.56$ to $0.21$~pp for patch occlusion,
and the largest single-condition effect falls from $22.5$~pp to at most
$2.5$~pp. The class-conditional spatial, spectral, and temporal structure
reported in the main text is therefore absent under shuffled labels and
cannot be attributed to the perturbation procedure itself.

\section{Computing Infrastructure and Software}
\label{sec:infra}

Experiments were conducted on two Linux GPU clusters. The main-table
results were produced primarily on an internal cluster equipped with Tesla
V100 (32~GB) GPUs (seven of the eight datasets), using Python 3.10, PyTorch
2.4.1 (CUDA 12.1, cuDNN 9.1), NumPy 2.1.2, and Zarr 2.18.3. The SleepEDF
main-table runs, all block-ablation runs, and the release validation were
executed on Volcano Engine ML Platform nodes equipped with
4$\times$NVIDIA H20 (96~GB) GPUs per node, with one training process pinned
to each GPU, using Python 3.11, PyTorch 2.8.0 (CUDA 12.8), and Zarr 2.18.7.
Per-run logs record the exact environment string (\texttt{torch/cuda/zarr}
versions), the resolved data root, the split manifest SHA-256, and a
configuration check line (\texttt{multi\_level\_readout=True}, selection
metric) for every training run.

\section{Independent Reproduction Check}
\label{sec:repro}

Before submission, the released code and configurations were re-executed
from scratch on a different hardware and software stack than the one used
for the primary results (H20 nodes with PyTorch 2.8.0/CUDA 12.8, versus
V100 with PyTorch 2.4.1/CUDA 12.1). Three checks were performed. First, the
released Full model builds to exactly 3{,}251{,}600 parameters under the
SEED configuration, identical to the original training code, and the Full
model together with all structurally distinct variant classes defined in
the \emph{Ablation Variant Definitions} section pass a forward/backward
smoke test with gradients flowing to every parameter. Second, an end-to-end re-run of BCIC2A and SHU (three
seeds each, full training protocol) reproduces the main-table mean
accuracies within the original three-seed standard-deviation bands:
$54.8$ versus $56.6 \pm 2.71$ on BCIC2A, and $61.0$ versus
$60.2 \pm 7.16$ on SHU. Third, every reported number is machine-audited
against the per-run result records
(\texttt{release\_audit/traceability\_audit.py}, deviations below $10^{-6}$).
Finally, the same cleaned release was re-executed on the \emph{original}
V100/PyTorch~2.4.1 stack for BCIC2A (three seeds, full protocol): the
per-split test accuracies ($53.5$/$58.3$/$58.0$, mean $56.6$) match the
original result records to the reported precision, confirming that the
release reproduces the primary results under the original environment and
that the cross-stack deviations described below are environmental.

Training is not run in deterministic mode. Across heterogeneous GPU
architectures and library versions, per-seed accuracies on small datasets
can vary across runs, because validation-based early
stopping amplifies kernel-level nondeterminism into different selected
checkpoints; the three-seed mean and its spread are stable. This is why all
results in this paper are reported as mean $\pm$ sample standard deviation
over three seeds rather than per-seed point values.

\section{Code Availability}
\label{sec:code}

The code repository for TriDimEEG is
\url{https://github.com/ncclab-sustech/TriDim_model}.
The release artifacts described in this supplement comprise the source
code, per-dataset YAML configurations, frozen split manifests with SHA-256
checksums, per-seed result records and training logs, and a
release-validation script (\texttt{scripts/validate\_release.py}).
Repository contents and licensing terms should be consulted before reuse.
The validation script checks the dataset registry, configuration
consistency, immutable manifests, and result-file completeness of the
release.

\section{Additional Architectural Controls}
\label{sec:controls}

Beyond the ablation variants reported in the main text, we evaluate two
additional architectural controls that isolate the contribution of the
tri-axis factorization itself. Both controls reuse the frozen
subject-level split manifests, the per-dataset configurations of
Supplementary Tables~\ref{tab:arch} and~\ref{tab:train}, and the identical training
protocol (seeds 5/42/43, validation-based checkpoint selection).

\paragraph{Criss-cross control.}
This control fixes the short-term axis $S$ as the embedding dimension and
applies only two attention operators, $\mathrm{Attn}_{c|s}$ over the
channel axis and $\mathrm{Attn}_{l|s}$ over the long-term axis, both with
$S$ as the embedding axis. The two operators share parameters, and their
outputs are combined by a fixed $0.5$ average in a direct residual
connection---there is no axis-role rotation and no learned attention view
fusion. The FFN path (three axis-specific MLPs with learned softmax
fusion) is identical to the Full model. The control uses $3{,}121{,}800$
parameters versus $3{,}246{,}609$ for the Full model under the same
configuration ($-3.84\%$), so the comparison is not confounded by
capacity.

Supplementary Table~\ref{tab:crisscross} reports the per-dataset results. The
criss-cross control falls below the Full model in accuracy on five of the
eight datasets; the exceptions are SHU-MI, SleepEDF, and PhysioNet-MI,
with relative improvements of approximately $7.6\%$, $0.3\%$, and $1.0\%$
over Full, respectively. Across all eight datasets, it reduces average
accuracy by approximately $3.1\%$ and average macro-F1 by $4.0\%$ relative
to Full. A
split-level paired Wilcoxon signed-rank test ($n = 8 \times 3 = 24$)
does not reach significance (accuracy $p = 0.143$; macro-F1 $p = 0.107$),
so we do not claim a significant per-pair degradation; the relevant
observation is the consistency of the gap across datasets rather than its
size on any single one.

\begin{table}[htbp]
\centering
\caption{Criss-cross control versus Full TriDimEEG: test accuracy and
macro-F1 (\%), reported as mean $\pm$ sample standard deviation across the
three subject-level splits (seeds 5/42/43).}
\label{tab:crisscross}
\setlength{\tabcolsep}{3pt}
\begin{tabular*}{\textwidth}{@{\extracolsep{\fill}}lcccc@{}}
\toprule
& \multicolumn{2}{c}{Accuracy} & \multicolumn{2}{c}{Macro-F1} \\
\cmidrule(lr){2-3}\cmidrule(lr){4-5}
Dataset & Full & criss-cross & Full & criss-cross \\
\midrule
AD65
& $63.8 \pm 6.10$
& $52.8 \pm 4.83$
& $60.4 \pm 7.62$
& $47.4 \pm 4.88$ \\

SleepEDF
& $85.6 \pm 2.19$
& $85.9 \pm 2.51$
& $66.9 \pm 3.17$
& $66.7 \pm 4.82$ \\

BCI-IV-2A
& $56.6 \pm 2.71$
& $54.5 \pm 8.34$
& $56.1 \pm 3.11$
& $54.2 \pm 8.58$ \\

SHU-MI
& $60.2 \pm 7.16$
& $64.8 \pm 7.40$
& $59.5 \pm 7.16$
& $64.3 \pm 7.17$ \\

PhysioNet-MI
& $60.4 \pm 8.00$
& $61.0 \pm 7.67$
& $60.6 \pm 7.83$
& $61.2 \pm 7.40$ \\

FACED
& $53.0 \pm 2.87$
& $52.0 \pm 2.60$
& $53.3 \pm 3.06$
& $52.4 \pm 2.51$ \\

SEED
& $52.9 \pm 2.22$
& $50.6 \pm 0.660$
& $51.2 \pm 1.95$
& $47.8 \pm 2.14$ \\

SEED-V
& $29.0 \pm 4.34$
& $25.5 \pm 2.53$
& $28.2 \pm 4.54$
& $24.2 \pm 3.92$ \\

\midrule
Average
& $\mathbf{57.7}$
& $55.9$
& $\mathbf{54.5}$
& $52.3$ \\
\bottomrule
\end{tabular*}
\end{table}

\clearpage
\paragraph{Flattened Transformer control.}
This control replaces the tri-axis encoder with a standard Transformer
operating on a flattened token sequence. Given the projected representation
$\mathbf{X}\in\mathbb{R}^{B\times C\times S\times L}$, the short-term and
long-term axes are flattened into a single sequence dimension, yielding
$\mathbf{X}_{\mathrm{flat}}\in\mathbb{R}^{B\times (S\cdot L)\times C}$.
Thus, each $(s,l)$ position forms one token, with the $C$-dimensional
channel representation serving as its embedding. The flattened sequence is
processed by the same number of standard pre-norm Transformer layers as
TriDimEEG, each consisting of multi-head self-attention followed by a
two-layer FFN with expansion ratio 2. The number of attention heads,
dropout rate, and LayerScale initialization are matched to the Full model.

For compatibility with the unchanged multi-level tri-axis readout, the
output of each Transformer layer is reshaped back to
$[B,C,S,L]$ before readout. All components outside the encoder---including
the frontend, tri-axis projections, multi-level tri-axis readout, and
classifier head---are kept identical to the Full model. As shown in
Supplementary Table~\ref{tab:flattf}, replacing the structure-preserving
TriDim encoder with this flattened-token Transformer reduces average
accuracy by approximately $1.2\%$ and average macro-F1 by $3.1\%$
relative to the Full model.

\begin{table}[!htbp]
\centering
\caption{Flattened Transformer control versus Full TriDimEEG: test accuracy and macro-F1 (\%),
reported as mean $\pm$ sample standard deviation across the three subject-level splits
(seeds 5/42/43).}
\label{tab:flattf}
\setlength{\tabcolsep}{4pt}
\begin{tabular*}{\textwidth}{@{\extracolsep{\fill}}lcccc@{}}
\toprule
& \multicolumn{2}{c}{Accuracy} & \multicolumn{2}{c}{Macro-F1} \\
\cmidrule(lr){2-3}\cmidrule(lr){4-5}
Dataset & Full & Flattened & Full & Flattened \\
\midrule
AD65
& $63.8 \pm 6.10$
& $59.6 \pm 12.4$
& $60.4 \pm 7.62$
& $51.1 \pm 16.7$ \\

SleepEDF
& $85.6 \pm 2.19$
& $84.8 \pm 1.92$
& $66.9 \pm 3.17$
& $66.0 \pm 3.93$ \\

BCI-IV-2A
& $56.6 \pm 2.71$
& $49.8 \pm 5.82$
& $56.1 \pm 3.11$
& $48.9 \pm 5.64$ \\

SHU-MI
& $60.2 \pm 7.16$
& $64.8 \pm 9.13$
& $59.5 \pm 7.16$
& $63.6 \pm 10.2$ \\

PhysioNet-MI
& $60.4 \pm 8.00$
& $61.1 \pm 6.55$
& $60.6 \pm 7.83$
& $60.9 \pm 6.28$ \\

FACED
& $53.0 \pm 2.87$
& $53.2 \pm 2.99$
& $53.3 \pm 3.06$
& $53.5 \pm 2.98$ \\

SEED
& $52.9 \pm 2.22$
& $55.5 \pm 2.54$
& $51.2 \pm 1.95$
& $55.1 \pm 3.12$ \\

SEED-V
& $29.0 \pm 4.34$
& $27.0 \pm 6.51$
& $28.2 \pm 4.54$
& $23.0 \pm 9.68$ \\

\midrule
Average
& $\mathbf{57.7}$
& $57.0$
& $\mathbf{54.5}$
& $52.8$ \\
\bottomrule
\end{tabular*}
\end{table}

\paragraph{An ordered evidence chain.}
Taken together with the structural ablations of the main text, the
architecture controls form an ordered evidence chain: Full TriDimEEG
($57.7$) $>$ sequential $C{\rightarrow}S{\rightarrow}L$ ($57.1$) $>$
Flattened Transformer ($57.0$) $>$ criss-cross ($55.9$) in average
accuracy. Removing the axis-role rotation while keeping two attention
operators (criss-cross) costs more than removing the tri-axis
factorization while keeping full self-attention (Flattened Transformer),
and both fall below the Full tri-axis block, supporting the view that the
structured three-axis decomposition---not merely the presence of
attention---underlies the cross-task robustness of the Full model.

\clearpage
\endgroup